\pdfoutput=1

\documentclass[11pt]{article}
\usepackage{amsmath}
\usepackage{algorithm}

\usepackage[final]{acl}

\usepackage{times}
\usepackage{dashrule}
\usepackage{arydshln}
\usepackage{latexsym}
\usepackage{booktabs}
\usepackage{multirow}
\usepackage{graphicx}
\usepackage[table]{xcolor}
\usepackage{colortbl}
\usepackage{array}  
\usepackage{tabularx}
\usepackage{subcaption}
\usepackage[T1]{fontenc}
\usepackage[utf8]{inputenc}
\usepackage{microtype}
\usepackage{inconsolata}
\usepackage{paralist}
\usepackage{longtable}
\usepackage{float}    
\usepackage{amssymb}
\usepackage{algorithmic}
\usepackage{fancyhdr}

\usepackage[export]{adjustbox}
\fancypagestyle{acceptedfooter}{
  \fancyhf{}
  \fancyfoot[C]{\footnotesize\itshape Accepted to Findings of the Association for Computational Linguistics: ACL 2026}

}

\title{Planning Beyond Text: Graph-based Reasoning for \\Complex Narrative Generation}

\author{Hanwen Gu$^{1,2,*}$ \quad Chao Guo$^{1,*,\dagger}$ \quad Junle Wang$^{2}$ \quad Wenda Xie$^{1}$ \quad Yisheng Lv$^{1,\dagger}$ \\
  $^{1}$Institute of Automation, Chinese Academy of Sciences \\
  $^{2}$Tencent Turing Lab \\
  \texttt{hanwengu@foxmail.com \quad chao.guo@ia.ac.cn \quad wangjunle@gmail.com} \\
  \texttt{xiewenda2024@ia.ac.cn \quad yisheng.lv@ia.ac.cn}
}

\begin{document}
\maketitle
\thispagestyle{acceptedfooter}
\begingroup
\renewcommand{\thefootnote}{\fnsymbol{footnote}}
\footnotetext[1]{These authors contributed equally.}
\footnotetext[2]{Corresponding authors.}
\endgroup
\setcounter{footnote}{0}
\begin{abstract}
While LLMs demonstrate remarkable fluency in narrative generation, existing methods struggle to maintain global narrative coherence, contextual logical consistency, and smooth character development, often producing monotonous scripts with structural fractures. 
To this end, we introduce PLOTTER, a framework that performs narrative planning on structural graph representations instead of the direct sequential text representations used in existing work.
Specifically, PLOTTER executes the Evaluate-Plan-Revise cycle on the event graph and character graph.
By diagnosing and repairing issues of the graph topology under rigorous logical constraints, the model optimizes the causality and narrative skeleton before complete context generation.
Experiments demonstrate that PLOTTER significantly outperforms representative baselines across diverse narrative scenarios. 
These findings verify that planning narratives on structural graph representations—rather than directly on text—is crucial to enhance the long context reasoning of LLMs in complex narrative generation.
\end{abstract}

\section{Introduction}
Although LLMs demonstrate remarkable proficiency in storytelling, long-form complex narrative generation remains challenging. It requires not only linguistic fluency but also the construction of a logically grounded narrative structure \citep{mirowski2023co,qian2024hollmwood,10085970}. To realize coordinated planning for long-form narratives, recent work has introduced iterative hierarchical planning \citep{yang2022re3,yang2023doc,xie2026plugandplaydramaturgedivideandconquerapproach} or role-playing strategies \citep{qian2024hollmwood,han2024ibsen} via LLM agents. 

However, these methods still struggle to maintain global narrative coherence, contextual logical consistency, and smooth character development, often producing monotonous scripts with structural fractures such as insufficient character motivation and a lack of necessary twists \citep{marco-etal-2024-pron,spangher-etal-2024-llms,wang-etal-2025-towards-novel}.
We argue the primary reason is that narrative planning directly on text representations is inefficient. Without explicit modeling of plot dependencies, such systems cannot effectively reason about the underlying cause-and-effect web or the evolving relationships among characters and events, ultimately limiting their ability to produce rigorous narrative structures \citep{sun2023event,zhang2024narrative}. 

To address this issue, we introduce \textbf{PLOTTER}, which stands for \textbf{PL}anning Bey\textbf{O}nd \textbf{T}ext: Graph-based Reasoning for Complex Narra\textbf{T}ive Gen\textbf{ER}ation. 
Unlike prior methods that plan on text directly, we plan narratives on graph structures, transforming script generation from a sequence planning problem into a dynamic graph generation and refinement problem.  
By applying the \emph{Evaluate-Plan-Revise} cycle onto the narrative topology, PLOTTER enables the model to diagnose and repair structural inconsistencies at the causal level before textual realization.

We represent narrative dynamics through two interacting structures: an \emph{Event Graph}, capturing the causal structure and skeleton of the plot, and a \emph{Character Graph}, modeling inter-character relationships. 
This design draws inspiration from classic narratological theories regarding action logic \citep{barthes1974sz} and character networks \citep{moretti2011network}. 
Consequently, the challenge of script writing is decomposed into atomic graph-editing operations—adding, deleting, or re-linking nodes and edges—to resolve causal gaps and character inconsistencies.

To enforce structural integrity, we deploy a graph-grounded refinement cycle with a multi-agent critique module to audit symbolic structures rather than raw text. These agents identify weaknesses—such as disconnected causal paths—and formulate revision strategies. A \emph{Constrained Graph Editor} then executes atomic operations to repair the narratives. 

We evaluate our framework across diverse narrative scenarios using multiple mainstream LLM backbones. Results show that our graph-based narrative reasoning approach significantly outperforms existing methods. 

Our main contributions are as follows:
\begin{itemize}
    \item We propose \textbf{PLOTTER}, the first framework to perform narrative planning on structural graph representations rather than the direct sequential text, enhancing long context causal reasoning in complex narrative generation. 
    \item We construct an LLM agent system for the iterative structural refinement of event and character graphs. By coordinating specialized agents to execute the \emph{Evaluate-Plan-Revise} cycle on the graph topology, we achieve precise structural diagnosis and repairs that are inaccessible to text-based narrative planners.

    \item Experiments show that \textbf{PLOTTER} significantly outperforms strong existing methods, verifying that elevating narrative planning from direct text representations to explicit graph representations is crucial for enhancing long-range reasoning in complex narratives.
\end{itemize}

\section{Related Work}

The landscape of automated narrative generation has evolved from sequence-to-sequence generation to LLM-based planning. We categorize existing literature into two streams: outline-based and role-play based narrative planning.

\subsection{Outline Based Narrative Planning}
To improve the coherence in long narrative contexts, most approaches adopt a ``Plan-and-Write'' paradigm and decompose the generation task into hierarchical stages guided by the planned outline. Frameworks like Re3 \citep{yang2022re3} utilize recursive prompting for context maintenance, while Detailed Outline Control (DOC) \citep{yang2023doc} and DOME \citep{wang2025dome} impose strict constraints to align generation with high-level outlines. Recent advancements such as CONCOCT \citep{wang2023pacing} further refine this by dynamically evaluating outline pacing. However, because these methods operate sequentially, logical inconsistencies in early steps often propagate downstream, causing cascading errors \citep{yang2022re3}. Furthermore, treating the outline as a rigid constraint restricts the flexibility required for complex revisions while avoiding extensive rewriting of preceding contexts \citep{yang2023doc}.

\subsection{Role-Play Based Narrative Planning}
Recent research has shifted focus from static planning to dynamic multi-agent simulation, where frameworks like HoLLMwood \citep{qian2024hollmwood}, Agents' Room \citep{huot2024agents}, IBSEN \citep{han2024ibsen}, and StoryWriter \citep{10.1145/3746252.3761616} assign specialized personas (e.g., Director, Actor) to distinct LLM instances. While these role-playing paradigms excel at stylistic diversity and dialogue richness, their coordination relies predominantly on unstructured natural language. Research indicates that such purely textual critique loops are susceptible to ambiguity and semantic drift over long contexts \citep{bae2024collective}. Consequently, without a shared symbolic state to ground collaboration, instructions from high-level planners are often misinterpreted by downstream agents, leading to hallucinations that contradict established narrative facts.

\subsection{Graph-Based LLM Reasoning}

Recent research integrates graph structures to enhance multi-hop reasoning for LLMs. For instance, Reasoning with Graphs (RwG) \citep{han-etal-2025-reasoning} structures implicit context into explicit graphs, Think-on-Graph (ToG) \citep{ICLR2024_10a6bdca} utilizes knowledge graphs for deep inference, and TG-LLM \citep{xiong2024large} employs temporal graphs to bolster chronological understanding. 
In the specific domain of LLM story generation,
recent work, such as R$^2$ \citep{li2024r2}, extracts graphs from a complete source text to build static external memories for generation reference. These works utilize graph structures as fixed contextual references for generation, while PLOTTER performs dynamic narrative planning via graph-structured representations of events and characters.

Unlike prior work that performs story planning directly on text representations, we use LLMs to plan narrative structure on graph-based representations. Specifically, we ensure logical and relational consistency through iterative generation and editing of event and character graphs via atomic edit operations, thereby improving plot coherence, engagement, and character development.

\begin{figure*}[t]
    \centering
    \includegraphics[width=\linewidth]{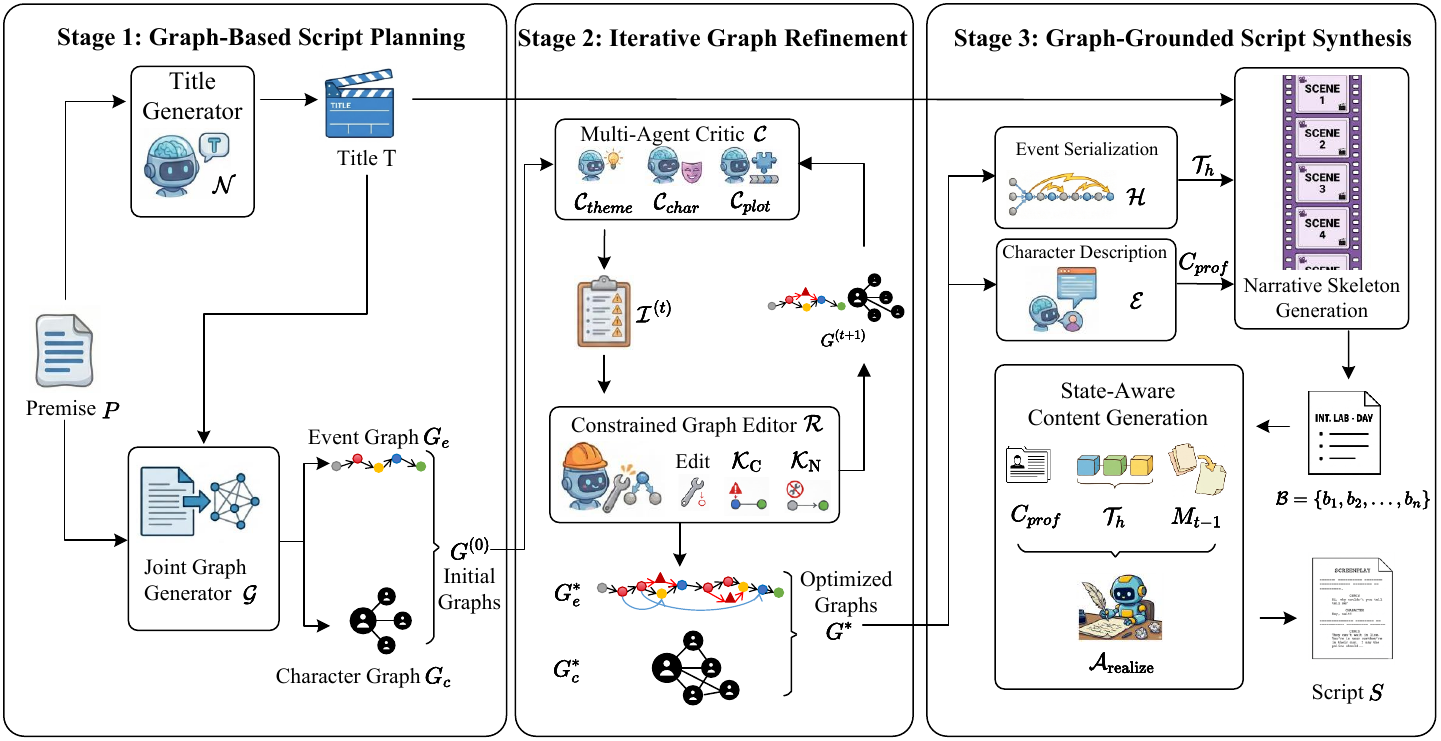}
    \caption{\textbf{Overview of the PLOTTER framework.} 
    (1) \textbf{Graph-Based Script Planning} initializes the narrative backbone comprising Event ($G_e$) and Character ($G_c$) graphs. 
    (2) \textbf{Iterative Graph Refinement} employs a Multi-Agent Critic ($\mathcal{C}$) to diagnose structural issues, which are resolved by a Constrained Graph Editor ($\mathcal{R}$) to produce an optimized graph ($G^*$). 
    (3) \textbf{Graph-Grounded Script Synthesis} serializes the graph via Event Serialization ($\mathcal{H}$) and Persona Expansion ($\mathcal{E}$), conditioning a State-Aware Generator to synthesize the final script ($S$).}
    \label{fig:framework}
\end{figure*}
\section{Method}
\label{sec:method}

\subsection{Task Definition}
Given a concise premise $P$, our goal is to generate a complete, high-quality script $S$. We propose a framework $\mathcal{F}$ that decomposes this complex generation task into three consecutive stages: (1) \textbf{Graph-based Narrative Planning}, which constructs the initial event backbone and character relation; (2) \textbf{Iterative Graph Refinement}, which optimizes the narrative (event and character) graph through an iterative \emph{Evaluate-Plan-Revise} cycle;
and (3) \textbf{Graph-Grounded Script Synthesis}, which transforms the structured graph into a natural language script. The overall framework is illustrated in Figure \ref{fig:framework}. 

\subsection{Stage 1: Graph-based Script Planning}
\label{ssec:rep}

First, a Title Generator $\mathcal{N}$ derives a thematic title from the premise $P$ (see Prompt \ref{prompt:title} in Appendix). Subsequently, a Joint Graph Generator $\mathcal{G}$ constructs the initial event and character graph representation, explicitly delineating both causal event dependencies and the web of character relationships (see Prompt \ref{prompt:story_graph} in Appendix):
\begin{equation}
    \qquad G^{(0)} = \mathcal{G}(\mathcal{N}(P), P)
\end{equation}
where $G^{(0)} = (G_e^{(0)}, G_c^{(0)})$ denotes the initial joint graph structure.

\paragraph{Event Graph Construction.}
Following Causal Network Theory \cite{trabasso1985causal}, the Event Graph $G_e = (V_e, E_e)$ functions as the causal skeleton. Each node $v_e \in V_e$ represents a distinct plot event with explicit attributes, including event ID, event description, narrative stage (e.g., \textit{Rising Action}, \textit{Climax}), and time index. The directed edges $E_e$ are not merely temporal links but encode narrative relation labels $\rho(e) \in \{\textsc{Causal}, \textsc{Foreshadowing}, \textsc{Suspense}\}$.
The Event Graph is generated through a structured prompt that instantiates these node attributes and edge relations under the above constraints (see Prompt \ref{prompt:story_graph} in Appendix).

\paragraph{Character Graph Construction.}
The Character Graph $G_c = (V_c, E_c)$ captures the sociodynamics of the narrative world. Each node $v_c$ encodes multi-dimensional persona slots---including core personality trait, internal conflict, external goal, and hidden secret---while edges $E_c$ represent evolving relationships with typed categories (Conflict, Cooperative, Emotional, Hidden) \cite{james1884art}. 
The Character Graph is generated simultaneously with the Event Graph using Prompt \ref{prompt:story_graph} in the Appendix, ensuring alignment between character relationships and plot events.

\subsection{Stage 2: Iterative Graph Refinement}
\label{sec:graph_optim}
Initial graphs generated by LLMs often exhibit structural issues or flat character arcs. 
To bridge the gap between a rough draft and a polished script, we introduce an iterative revision mechanism on narrative graphs that mimics the \emph{Evaluate-Plan-Revise} cognitive process of expert writers \cite{flower1981cognitive}. 
The process alternates between a multi-agent critique module $\mathcal{C}$ and a graph refinement module $\mathcal{R}$. We define the set of critics as $\mathcal{C} = \{\mathcal{C}_{\text{theme}}, \mathcal{C}_{\text{char}}, \mathcal{C}_{\text{plot}}\}$, representing specialized agents that evaluate thematic consistency, character depth, and plot logic, respectively. The graph update at iteration $t$ is formulated as:
\begin{equation}
    G^{(t+1)} = \mathcal{R}\big(\mathcal{C}(G^{(t)}),\, G^{(t)}\big)
\end{equation}
where $G^{(t)}$ denotes the graph state at step $t$, and $\mathcal{C}(G^{(t)})$ represents the aggregated feedback used by $\mathcal{R}$ to execute topological modifications.

The complete procedure is formalized in Algorithm~\ref{alg:refinement}.

\paragraph{Narrative Critics ($\mathcal{C}$).}
We deploy a suite of specialized agents that evaluate the narrative graph through a hierarchical critique process. Reflecting the fundamental principles of story craft \cite{mckee1997story,egri1946art}, the agents execute in a fixed sequence to ensure each narrative dimension builds upon a solid foundation:

\begin{enumerate}
    \item \textbf{Theme Critic ($\mathcal{C}_{\text{theme}}$)} identifies instances where the storyline drifts from its core message or where the theme is merely stated through exposition rather than being "shown" through conflict and symbolism.
    
    \item \textbf{Character Critic ($\mathcal{C}_{\text{char}}$)} builds upon thematic coherence to audit persona depth. It diagnoses flat development where characters lack growth, flags decisions missing clear internal or external motivation, and detects sudden attitude shifts that lack the necessary psychological buildup.
    
    \item \textbf{Plot Critic ($\mathcal{C}_{\text{plot}}$)} integrates the preceding elements into a causally sound structure. It audits structural integrity by detecting causal discontinuities and logical contradictions. Furthermore, it ensures narrative engagement by flagging missing foreshadowing and monotone plotlines lacking turning points.
\end{enumerate}

The analysis function of each agent $\mathcal{C}_i$ generates a structured issue list, formally denoted as $\mathcal{I}_i$. 
Cross-agent validation is further applied. Only edits that receive consistent support across agents proceed to execution, which limits the propagation of local reasoning errors.
This list comprises five key components as defined in Table \ref{tab:issue_structure}. The complete taxonomy and details of all issue types are presented in Table \ref{tab:critic}. The specific prompts used for issue identification are detailed in Prompts \ref{prompt:theme_agent}--\ref{prompt:plot_agent}.

\paragraph{Constrained Graph Editor ($\mathcal{R}$).}
Upon receiving the issue list $\mathcal{I}$, the Editor $\mathcal{R}$ resolves structural deficiencies by mapping each issue $i \in \mathcal{I}$ to a sequence of atomic edit operations $\omega \in \Omega$ (e.g., \texttt{Add-Plot-Bridge}, \texttt{Revise-Event}; see Table \ref{tab:editor} for full definitions). To ensure these modifications do not introduce new contradictions and mitigate error accumulation, the Editor operates under strict verification. For any proposed edit $\omega$, the resulting graph must satisfy the core narrative constraints:

\begin{equation}
    \text{Edit}(G, \omega) \models \mathcal{K}_{\text{C}} \land \mathcal{K}_{\text{N}}
\end{equation}

The edits are grounded by two deterministic symbolic constraints:
\begin{itemize}
    \item \textbf{Causal Rationality ($\mathcal{K}_{\text{C}}$):} We enforce that the causal subgraph must remain a Directed Acyclic Graph (DAG). This mathematically guarantees the forward flow of time and prevents logical loops, ensuring that every narrative event is preceded by a valid cause.
    \item \textbf{Narrative Completeness ($\mathcal{K}_{\text{N}}$):} We verify that every node remains reachable from the \textit{Beginning} and maintains a path to the \textit{Ending}. This ensures that all events are logically integrated into the main storyline and prevents the existence of isolated nodes that do not contribute to the global narrative arc.
\end{itemize}

If a modification violates either axiom, it is rejected, preserving the topological validity of the plot. The final optimized graph is denoted as $G^* = (G_e^*, G_c^*)$. Since the symbolic constraints ($\mathcal{K}_{\text{C}}$ and $\mathcal{K}_{\text{N}}$) are verified deterministically without LLM involvement, structurally invalid edits cannot propagate. 

\begin{algorithm}[h]
\small 
\renewcommand{\algorithmicrequire}{\textbf{Input:}}
\renewcommand{\algorithmicensure}{\textbf{Output:}}
\caption{Hierarchical Iterative Graph Refinement}
\label{alg:refinement}
\begin{algorithmic}[1]
\REQUIRE Initial graph $G^{(0)}$, Critics $\mathcal{C}$, Editor $\mathcal{R}$, Max iterations $K$
\ENSURE Optimized narrative graph $G^*$
\STATE $G^* \leftarrow G^{(0)}$ 
\FOR{$t = 1$ \TO $K$}
    \STATE $\textit{Updated} \leftarrow \FALSE$
    \FOR{\textbf{each} Agent $\mathcal{C}_i \in \mathcal{C}$} 
        \STATE $\mathcal{I}_i \leftarrow \mathcal{C}_i(G^*)$ \COMMENT{Detect structural issues}
        \IF{$\mathcal{I}_i \neq \emptyset$}
            \STATE $\omega \leftarrow \text{GenerateOps}(\mathcal{I}_i)$ \COMMENT{Map issues to $\omega$}
            \STATE $G' \leftarrow \text{Edit}(G^*, \omega)$ 
            \STATE $\textit{Updated} \leftarrow \TRUE$
            \IF{$G' \models \mathcal{K}_{\text{C}} \land \mathcal{K}_{\text{N}}$}
                \STATE $G^* \leftarrow G'$ \COMMENT{Accept valid edits}
            \ENDIF
        \ENDIF
    \ENDFOR
    \IF{$\textit{Updated} = \FALSE$}
        \STATE \textbf{break}
    \ENDIF
\ENDFOR
\RETURN $G^*$
\end{algorithmic}
\end{algorithm}

\subsection{Stage 3: Graph-Grounded Script Synthesis}
\label{ssec:gen}

In the final stage, we transform the optimized symbolic graphs $G^*$ into a coherent textual script $S$. To bridge the gap between graph structures and linear text generation, we employ a strategy of \textit{Graph Serialization} followed by \textit{State-Aware Generation}.

\subsubsection{Graph Serialization}
Directly feeding raw graph definitions to an LLM often obfuscates narrative temporal dependencies. We therefore serialize symbolic graphs into a logically ordered textual sequence that preserves topology, so the generator can consume structural constraints without losing causal or relational context.

\paragraph{Event Serialization.}
We serialize the Event Graph $G_e^*$ into a hierarchical event plan $\mathcal{T}_h=\mathcal{H}(G_e^*)$ via a deterministic depth-first traversal on the graph induced by causal and suspense relations. 
$\mathcal{T}_h$ specifies event-level progression and relation constraints. 
The traversal starts from events with no in-degree, visits each event once, and prioritizes suspense successors over causal successors when multiple successors are eligible; 
ties within the same relation type are resolved by the original chronological order in $G_e^*$. Foreshadowing edges are not included in the traversal and are instead retained as cross-event cues in the serialized output. 

\paragraph{Character Description.}
Simultaneously, we expand the concise nodes in $G_c^*$ into detailed character profiles $C_{prof} = \mathcal{E}(G_c^*)$ (see Prompt \ref{prompt:character_expansion} in Appendix). This step fleshes out hidden backstories, providing semantic anchors for subsequent dialogue generation.

\subsubsection{Progressive Script Generation}
With the serialized events $\mathcal{T}_h$ and character profiles $C_{prof}$, we generate script components progressively. In this setup, $\mathcal{T}_h$ provides event-level structural constraints, while scene beats provide scene-level realization units.

\paragraph{Narrative Skeleton Generation.}
To ensure global consistency and thematic continuity before drafting detailed dialogue, we first generate a comprehensive sequence of scene beats $\mathcal{B}$ in a single pass (see Prompt \ref{prompt:scene_batch} in Appendix), denoted as $\mathcal{B} = \{b_1, b_2, \dots, b_n\}$. This generation process is formulated as:
\begin{equation}
    \mathcal{B} = \text{LLM}(\mathcal{T}_h, C_{\text{prof}}, T)
\end{equation}
Each generated beat $b_i$ encapsulates a slugline, specific plot points, and pivotal character moments, strictly adhering to the causal and temporal dependencies established in the structural graph. Concretely, beats are generated conditionally on $\mathcal{T}_h$, so event structure is preserved while local scene content remains flexible.

\paragraph{State-Aware Content Generation.}
Finally, we flesh out each beat $b_i$ into a full script scene $\sigma_i$ through State-Aware Content Generation $\mathcal{A}_{\text{realize}}$. To maintain cross-scene coherence, the generator maintains a dynamic narrative state $M_i$. For each scene $i$, the generation is conditioned on:
\begin{itemize}
\item \textbf{Event Relation:} The relational context extracted from $\mathcal{T}_h$. By referencing the specific edge types (e.g., \textit{Suspense} or \textit{Conflict}) connected to the current event, the model aligns the scene's emotional tone and causal logic with the pre-optimized graph structure.
\item \textbf{Character Persona:} The comprehensive profiles $C_{prof}$ derived from the Character Graph. These serve as semantic anchors, ensuring that dialogue and actions remain consistent with the speaker's established voice and motivations.
\item \textbf{Contextual Memory:} The rolling narrative state $M_{i-1}$ that tracks the evolving history. $M$ is updated iteratively to incorporate previous dialogue and plot developments, ensuring referential continuity and preventing logical drift during long-range generation.
\end{itemize}
Formally, the transition from symbolic structure to script is performed by the state-aware realization function $\mathcal{A}_{\text{realize}}$:
\begin{equation}
    \sigma_i = \mathcal{A}_{\text{realize}}(b_i, \mathcal{T}_h, C_{prof}, M_{i-1})
\end{equation}
where $M_0 = \emptyset$. The final script $\mathcal{S}$ is produced by the ordered concatenation of all realized scenes: $\mathcal{S} = \bigcup_{i=1}^n \sigma_i$.
The complete generation procedure is summarized in Algorithm~\ref{alg:realization}.

\begin{algorithm}[!t]
\small 
\renewcommand{\algorithmicrequire}{\textbf{Input:}}
\renewcommand{\algorithmicensure}{\textbf{Output:}}
\caption{State-Aware Content Generation}
\label{alg:realization}
\begin{algorithmic}[1]
\REQUIRE Narrative Skeleton $\mathcal{B} = \{b_1, \dots, b_n\}$, Serialized Events $\mathcal{T}_h$, Character Profiles $C_{prof}$
\ENSURE Final Textual Script $\mathcal{S}$
\STATE $\mathcal{S} \leftarrow \emptyset$
\STATE $M_0 \leftarrow \emptyset$ \hfill \COMMENT{Initialize Contextual Memory}

\FOR{$i = 1$ \TO $n$}
    \STATE \COMMENT{\textit{1. State-Aware Realization}}
    \STATE $\sigma_i \leftarrow \mathcal{A}_{\text{realize}}(b_i, \mathcal{T}_h, C_{prof}, M_{i-1})$ \hfill \COMMENT{Generate scene conditioned on history}
    
    \STATE \COMMENT{\textit{2. Narrative State Tracking}}
    \STATE $M_i \leftarrow \text{Summarize}(\sigma_1, \dots, \sigma_i)$ \hfill \COMMENT{Update $M_i$ with latest developments}
    \STATE $\mathcal{S} \leftarrow \mathcal{S} \cup \{ \sigma_i \}$
\ENDFOR
\RETURN $\mathcal{S} = (\sigma_1, \sigma_2, \dots, \sigma_n)$
\end{algorithmic}
\end{algorithm}

\section{Experiment}
\subsection{Dataset}
To provide a rigorous evaluation with a broad coverage, we construct a dataset that balances diversity with complexity. While prior studies in narrative generation~\cite{yang2022re3,yang2023doc,mirowski2023co,qian2024hollmwood} typically use 20--60 LLM-generated premises as an evaluation dataset, we curate 50 premises from a hybrid mix of high-quality sources. Specifically, we sample from 3 human-curated datasets (MoPS~\cite{ma-etal-2024-mops}, WritingPrompts~\cite{fan-etal-2018-hierarchical}, ROCStories~\cite{mostafazadeh-etal-2016-corpus}) and 2 LLM-generated sources (DOC~\cite{yang2023doc}, GPT-4.1), as detailed in Appendix~\ref{sec:dataset_details}. These premises span 9 genres (Sci-Fi, Drama, Crime, Thriller, Fantasy, Romance, Horror, General, and Other) with a mean Type--Token Ratio (TTR) of 0.80, indicating high lexical diversity across the evaluation set. This configuration provides a broader distribution than those used in most prior work and a more robust evaluation. 

\begin{table*}[ht]
\centering
\setlength\tabcolsep{4pt}
\fontsize{7}{7}\selectfont
\begin{tabular}{clcccccccccc}
\toprule[1.3pt]
\multirow{2}{*}{Backbone} & \multirow{2}{*}{Method} & \multicolumn{2}{c}{Narrative $\uparrow$} & \multicolumn{2}{c}{ThematicExpression $\uparrow$} & \multicolumn{2}{c}{Characterization $\uparrow$} & \multicolumn{2}{c}{DramaticEngagement $\uparrow$} & \multicolumn{2}{c}{PremiseFidelity $\uparrow$} \\
\cmidrule(lr){3-4} \cmidrule(lr){5-6} \cmidrule(lr){7-8} \cmidrule(lr){9-10} \cmidrule(lr){11-12}
& & Storyline  & Script & Storyline  & Script & Storyline  & Script & Storyline  & Script & Storyline  & Script \\
\midrule

\multirow{9}{*}{\centering GPT4.1}
& \textit{\quad LLM Plan and Write wins} & 6 & 28 & 0 & 0 & 0 & 0 & 0 & 4 & 18 & 18 \\
& \textbf{\quad Ours wins}               & \textbf{94} & \textbf{72} & \textbf{100} & \textbf{100} & \textbf{100} & \textbf{100} & \textbf{100} & \textbf{96} & \textbf{34} & \textbf{40} \\
& \quad ties                             & 0 & 0 & 0 & 0 & 0 & 0 & 0 & 0 & 48 & 42 \\
& \textit{\quad Dramatron wins}          & 0 & 16 & 0 & 10 & 0 & 20 & 2 & 28 & 0 & 0 \\
& \textbf{\quad Ours wins}               & \textbf{100} & \textbf{74} & \textbf{100} & \textbf{90} & \textbf{100} & \textbf{76} & \textbf{98} & \textbf{72} & 2 & \textbf{14} \\
& \quad ties                             & 0 & 1 & 0 & 0 & 0 & 4 & 0 & 0 & 98 & 86 \\
& \textit{\quad DOC wins}                & 38 & 8 & 14 & 14 & 10 & 8 & 34 & 8 & 16 & 16 \\
& \textbf{\quad Ours wins}               & \textbf{62} & \textbf{92} & \textbf{86} & \textbf{86} & \textbf{90} & \textbf{92} & \textbf{66} & \textbf{92} & 10 & \textbf{44} \\
& \quad ties                             & 0 & 0 & 0 & 0 & 0 & 0 & 0 & 0 & 74 & 40 \\
\midrule

\multirow{9}{*}{\centering DeepSeek R1}
& \textit{\quad LLM Plan and Write wins} & 6 & 6 & 0 & 0 & 2 & 0 & 2 & 2 & 2 & 2 \\
& \textbf{\quad Ours wins}               & \textbf{94} & \textbf{94} & \textbf{100} & \textbf{100} & \textbf{98} & \textbf{100} & \textbf{98} & \textbf{98} & \textbf{94} & \textbf{94} \\
& \quad ties                             & 0 & 0 & 0 & 0 & 0 & 0 & 0 & 0 & 4 & 4 \\
& \textit{\quad Dramatron wins}          & 44 & 52 & 18 & 50 & 38 & 42 & 48 & 46 & 4 & 4 \\
& \textbf{\quad Ours wins}               & 42 & 48 & \textbf{82} & \textbf{50} & \textbf{60} & \textbf{58} & \textbf{52} & \textbf{54} & 4 & \textbf{8} \\
& \quad ties                             & 14 & 0 & 0 & 0 & 2 & 0 & 0 & 0 & 92 & 88 \\
& \textit{\quad DOC wins}                & 48 & 16 & 40 & 24 & 28 & 14 & 46 & 14 & 30 & 32 \\
& \textbf{\quad Ours wins}               & \textbf{52} & \textbf{84} & \textbf{60} & \textbf{76} & \textbf{72} & \textbf{86} & \textbf{54} & \textbf{86} & 18 & \textbf{48} \\
& \quad ties                             & 0 & 0 & 0 & 0 & 0 & 0 & 0 & 0 & 52 & 20 \\
\midrule

\multirow{9}{*}{\centering Qwen3}
& \textit{\quad LLM Plan and Write wins} & 14 & 12 & 0 & 0 & 0 & 0 & 8 & 4 & 12 & 14 \\
& \textbf{\quad Ours wins}               & \textbf{86} & \textbf{88} & \textbf{100} & \textbf{100} & \textbf{100} & \textbf{100} & \textbf{92} & \textbf{96} & \textbf{66} & \textbf{68} \\
& \quad ties                             & 0 & 0 & 0 & 0 & 0 & 0 & 0 & 0 & 22 & 18 \\
& \textit{\quad Dramatron wins}          & 32 & 36 & 8 & 16 & 8 & 30 & 28 & 40 & 8 & 2 \\
& \textbf{\quad Ours wins}               & \textbf{64} & \textbf{64} & \textbf{92} & \textbf{84} & \textbf{92} & \textbf{70} & \textbf{70} & \textbf{60} & 8 & \textbf{16} \\
& \quad ties                             & 0 & 0 & 0 & 0 & 0 & 0 & 2 & 0 & 84 & 82 \\
& \textit{\quad DOC wins}                & 30 & 22 & 26 & 24 & 20 & 20 & 28 & 22 & 34 & 26 \\
& \textbf{\quad Ours wins}               & \textbf{70} & \textbf{78} & \textbf{74} & \textbf{76} & \textbf{80} & \textbf{80} & \textbf{72} & \textbf{78} & 22 & \textbf{60} \\
& \quad ties                             & 0 & 0 & 0 & 0 & 0 & 0 & 0 & 0 & 44 & 14 \\
\bottomrule[1.3pt]
\end{tabular}
\caption{Pairwise comparison of storyline and overall script between our method and baselines under five evaluation dimensions. Experiments are conducted using GPT-4.1, DeepSeek R1, and Qwen3 as backbone models. Results demonstrate the superiority of our method in narrative coherence, thematic engagement, and character development.}
\label{tab:comparison}
\end{table*}

\subsection{Baselines}
We evaluate our method against three representative state-of-the-art script generation frameworks: \textbf{LLM-Plan-Write}, \textbf{Dramatron}, and \textbf{DOC}. Details of these baselines are provided in Appendix \ref{appendix:baselines}.

To ensure a fair comparison, we implement all methods using 3 mainstream LLM backbones respectively: \textbf{GPT-4.1} \cite{openai2025gpt41}, \textbf{DeepSeek-R1} (\texttt{deepseek-r1-250120})~\cite{deepseekai2025deepseekr1incentivizingreasoningcapability}, and \textbf{Qwen3} (\texttt{qwen3-235b-a22b})~\cite{yang2025qwen3technicalreport}. Premises are kept consistent across all methods and backbones. 
This multi-backbone setting ensures that the performance evaluations are not skewed by the inherent biases of a specific model family, nor by the self-preference bias caused by homologous generation and evaluation models.

\subsection{Evaluation Method}
Following prior work, we adopt a pairwise comparison using GPT-4.1 as the evaluator and carefully designed bias-mitigation strategies for LLM-based assessment, at both storyline and full-script levels (see Appendix~\ref{sec:evaluation_details} for details). In addition, human evaluation corroborates the validity of these results. The evaluator selects the better script or reports a tie across five dimensions: Narrative, Thematic Expression, Characterization, Dramatic Engagement, and Premise Fidelity (see Table~\ref{tab:evaluation_dimensions} for details). These dimensions are derived from classical dramatic theory \cite{mckee1997story} and align with established metrics in mainstream script generation work \cite{yang2022re3,yang2023doc,mirowski2023co,qian2024hollmwood}.

\subsection{Quantitative Results}

\renewcommand{\arraystretch}{1.2}

\paragraph{LLM Evaluation.} 
As shown in Table~\ref{tab:comparison}, PLOTTER consistently outperforms baselines across five evaluation dimensions. The most dominant gains appear in Narrative, Thematic Expression, Characterization, and Dramatic Engagement. These results directly validate the effectiveness of our Multi-Agent Critique and Constrained Graph Editor. Unlike baselines that rely on linear text generation, our method establishes a structurally valid narrative skeleton with the event graph before script text is produced, which improves long-range coherence and quality of the Narrative
and Thematic Expression.

The performance gap highlights the fundamental difference between our graph-based reasoning and existing methods. LLM Plan-and-Write and Dramatron suffer from a lack of structured refinement, and even DOC—which uses an iterative approach—is limited by a static, hierarchical text outline. By employing an \emph{Evaluate-Plan-Revise} cycle that operates directly on graph nodes and edges, our method identifies and resolves character inconsistencies and dramatic lulls more effectively than text-based iteration. This results in significantly higher win rates in Characterization and Dramatic Engagement, proving that planning on a graph structure is inherently superior to editing traditional linear outlines for complex storytelling.

\paragraph{Human Evaluation.}
Beyond LLM-based evaluation, we further conducted a human study to verify the reliability of LLM evaluation. The results show strong inter-rater agreement (Fleiss' $\kappa$ = 0.688) and alignment with the LLM evaluation (Cohen’s $\kappa$ = 0.834).
(Details in Appendix~\ref{sec:appendix_human}).

\paragraph{Objective Metrics.}
Due to the absence of a unified and comprehensive evaluation metric, existing methods widely rely on LLM and human evaluation. To mitigate potential length-caused confounds and bias, we report the statistical length of generated narratives and formula-based quantitative metrics for information density and redundancy in Table~\ref{tab:auto_metrics}. All methods produce scripts of comparable length ($\sim$10k words), confirming that our improvements are not attributable to length inflation. PLOTTER achieves the highest Distinct-2 and MATTR scores and the lowest Self-BLEU, indicating that its advantage stems from substantive content diversity rather than superficial lexical repetition.
To further assess stability across narrative types, we additionally analyze cross-genre variation and edge-pattern consistency, showing that performance remains stable across the nine genres in our dataset (see Appendix~\ref{sec:appendix_genre}).
\begin{table}[h]
\centering
\small
\renewcommand{\arraystretch}{1.1}
\resizebox{\columnwidth}{!}{
\begin{tabular}{lcccc}
\toprule
\textbf{Method} & \textbf{Words} & \textbf{Distinct-2}$\uparrow$ & \textbf{MATTR}$\uparrow$ & \textbf{Self-BLEU}$\downarrow$ \\
\midrule
PLOTTER         & 9738\scriptsize{$\pm$593}  & \textbf{0.793} & \textbf{82.6} & \textbf{0.017} \\
Dramatron       & 9792\scriptsize{$\pm$775}  & 0.778 & 81.4 & 0.022 \\
DOC             & 10330\scriptsize{$\pm$618} & 0.680 & 65.4 & 0.090 \\
LLM-Plan-Write  & 9683\scriptsize{$\pm$577}  & 0.752 & 79.2 & 0.031 \\
\bottomrule
\end{tabular}}
\caption{Objective evaluation metrics.}
\label{tab:auto_metrics}
\end{table}

\subsection{Qualitative Results}
\label{sec:qualitative}
To demonstrate how PLOTTER improves narrative structures, we present case studies on agent operations, end-to-end graph evolution, and comparative evaluation examples. 

\paragraph{Agent Operations}
In the presented example, the protagonist, Elias, undergoes a psychological transition from defeat to counter-attack. The initial generation exhibited a severe logical violation: the narrative jumped directly from Scene 7 (The Defeat) to Scene 8 (The Climax) without sufficient build-up.

\begin{itemize}
\item \textbf{Issue 1: Motive-Weak.} As illustrated in Figure~\ref{fig:case_topology}, the Character Critic flagged a Motive-Weak issue between Event 7 and Event 8, indicating an abrupt psychological transition. To repair this, the Constrained Graph Editor executed an Add-Plot-Bridge operation. By querying the \textit{Event Graph} for context anchors, the system retrieved the ``Missing People'' motif from Scene 1 and generated a bridging node where Elias's internal monologue converts guilt into the resolve required for the climax.

\item \textbf{Issue 2: Discontinuity.} A single bridging node is often insufficient for complex transitions. As shown in Figure~\ref{fig:case_structural}, a Discontinuity was detected where Elias's sudden leadership felt unearned. To restore causal sufficiency, the editor applied a tri-partite Add-Plot-Bridge strategy (``Trinity of Action''):
(1) Why: an internal node where Elias accepts his past arrogance.
(2) Who: a social node where Elias reconciles with his partner Leilani.
(3) How: a tactical node establishes the capability to mobilize the crowd.
This multi-node insertion shows that resolving narrative discontinuity requires constructing robust causal chains rather than simple text smoothing.
\end{itemize}

\paragraph{Graph Evolution.}
We present a representative evolution from the initial to the final graphs in Section~\ref{sec:toy_example}. Figure~\ref{fig:event_graph} makes the edit process explicit by tracing how Critic diagnoses are resolved through concrete Editor operations.

\paragraph{Comparative Evaluation Examples.}
Table~\ref{tab:script_excerpts} presents a comparative case study of the generated content against the baseline, while Table~\ref{tab:case_comparison} provides a comparison of their respective LLM evaluations. The evaluator prefers PLOTTER on Narrative, Thematic Expression, Characterization, and Dramatic Engagement. 

We also provide a statistical analysis of cross-genre generalization in Appendix~\ref{sec:appendix_genre}. 
Performance remains stable across genres, and agentic operations remain consistent, indicating that the observed gains are not limited to a single representative scenario.

\subsection{Ablation Study}
\label{sec:ablation_main}

To systematically verify the efficacy of \textbf{PLOTTER}, we conducted a comprehensive ablation study. 
We compare the Full Module against two categories of variants:
\begin{itemize}
    \item \textbf{Ablated Variants} (\texttt{w/o Theme}, \texttt{w/o Character}, \texttt{w/o Plot}): The corresponding agent is deactivated to verify the necessity of each reasoning dimension.
    \item \textbf{Isolated Variants} (\texttt{Theme Only}, \texttt{Character Only}, \texttt{Plot Only}): Only one dimension agent is active, prohibiting cross-dimensional feedback, to test the synergy of the modules in improving the narrative graph.
\end{itemize}

\begin{figure}[h]
    \centering
    \includegraphics[width=0.95\linewidth]{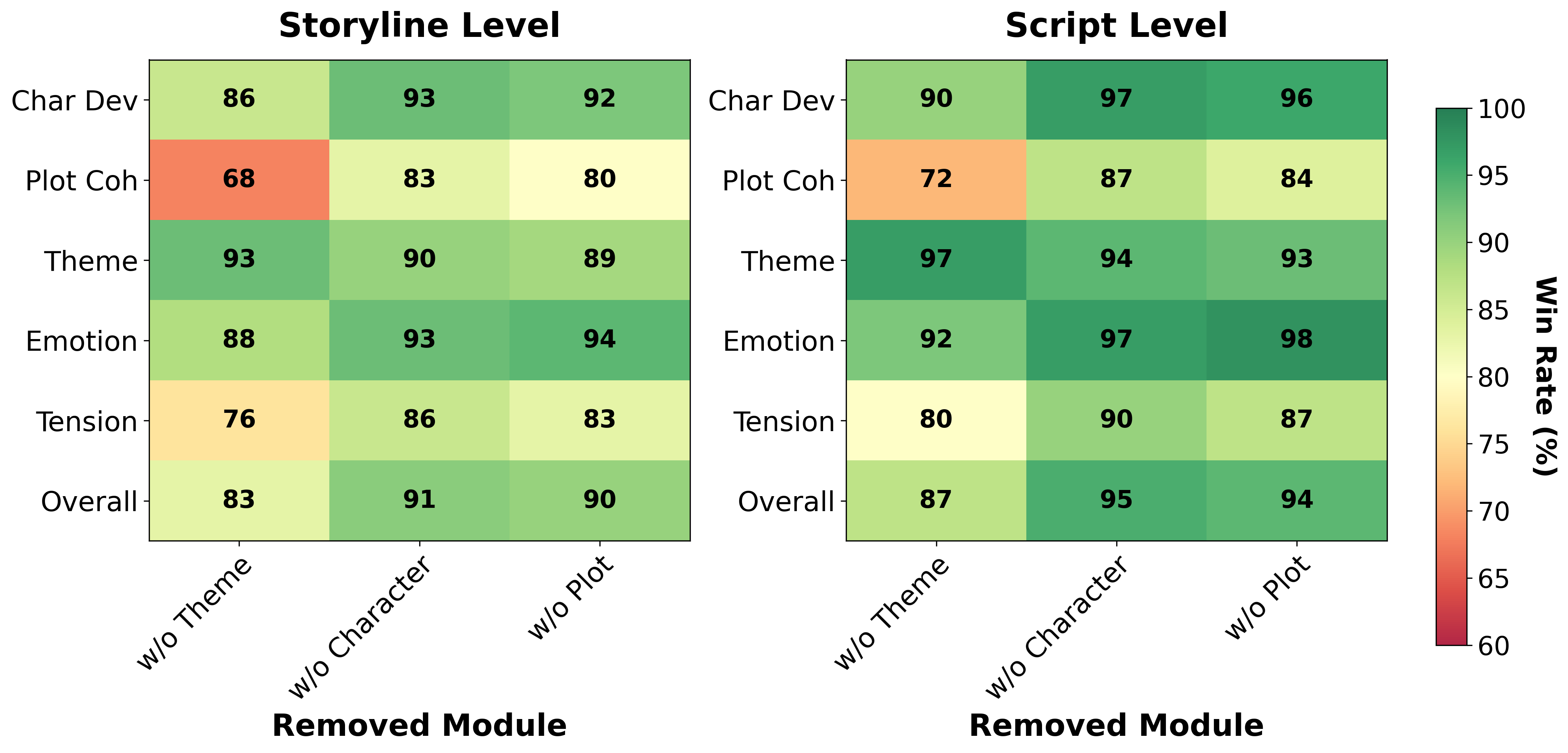}
    \caption{\textbf{Module Necessity Analysis.} Each cell shows the win rate (\%) of the full model against the corresponding ablated variant. Every reasoning agent within the narrative graph is critical for the holistic quality.}
    \label{fig:heatmap}
\end{figure}

\paragraph{Module Necessity.}
As illustrated in Figure~\ref{fig:heatmap}, each cell reports the win rate of the full PLOTTER against an ablated variant. 
The full module consistently outperforms all ablated variants across all evaluation dimensions. 
Removing the Character or Plot module yields the largest performance collapse, while removing the Theme module shows comparatively smaller overall drop but a clear disadvantage in thematic quality. It shows that the collaborative
reasoning between character psychology
and plot structure is essential for achieving high-quality character development and emotional resonance.
This confirms that every reasoning agent
within the narrative graph is critical for the
holistic quality of the generated script.

\paragraph{The ``$1+1>2$'' Synergy Effect.}

To verify inter-module coordination, we compared the Full Module against single-agent variants (Character, Plot, or Theme) in Stage 2 by measuring their respective win rates against the \textit{w/o Stage 2} baseline. 

\begin{figure}[h]
    \centering
    \includegraphics[width=0.85\linewidth]{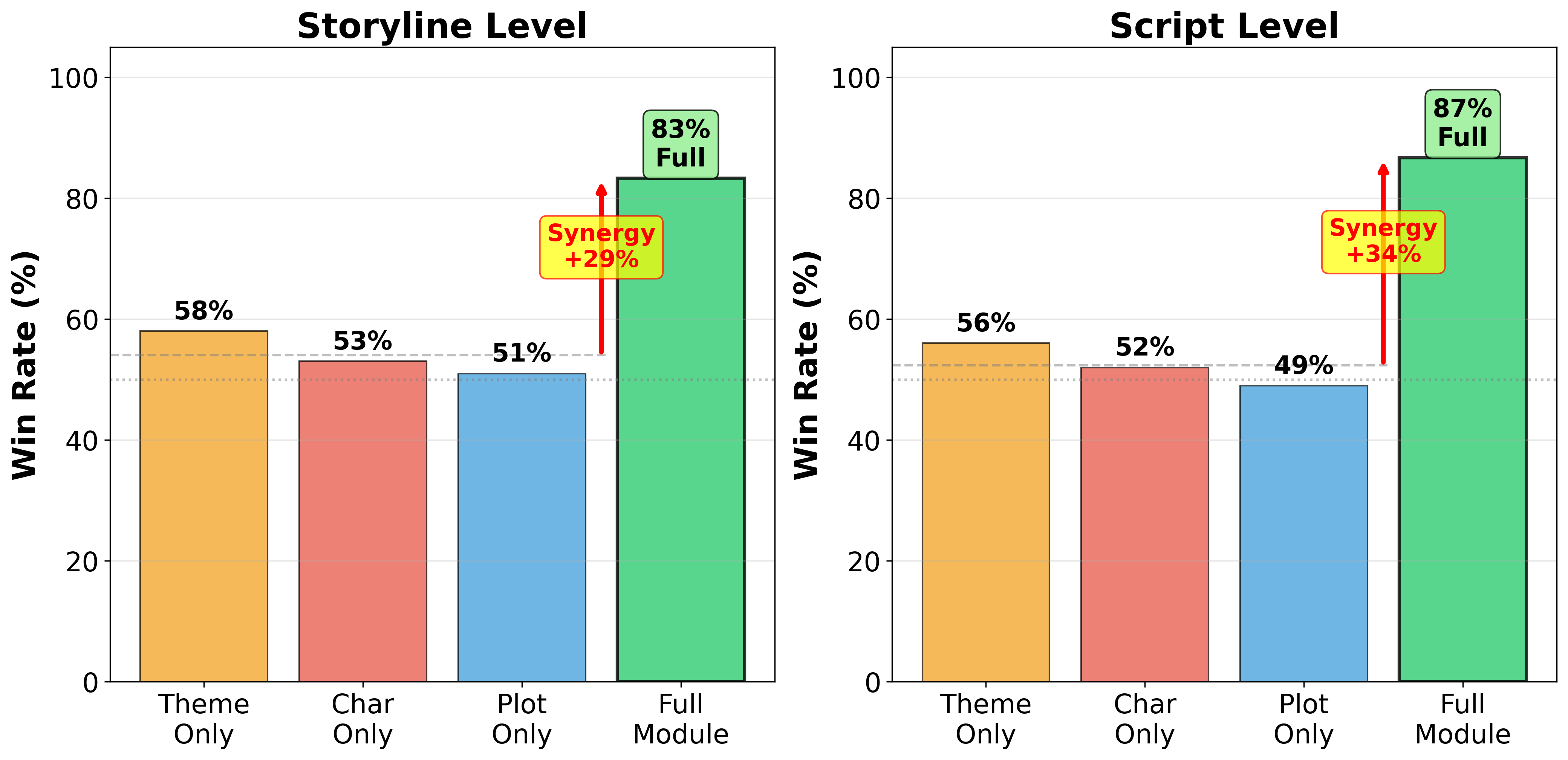}
    \caption{\textbf{Synergy Effect Analysis.} The incremental gains from individual modules sum to significantly less than the total performance improvement, indicating the synergy effect of modules during Stage 2 narrative graph refinement.}
    \label{fig:synergy_waterfall}
\end{figure}

As shown in Figure~\ref{fig:synergy_waterfall}, single-module variants yield only marginal improvements relative to \textit{w/o Stage 2}, whereas the full model achieves a dominant advantage (win rate $>$80\%). 
Correspondingly, the full model yields sizable synergy gaps of +29\% at the Storyline level and +34\% at the Script level over the average win rate of single-module variants. 
This massive jump significantly exceeds the linear sum of individual gains, indicating that joint optimization across modules is substantially more effective than activating any single module in isolation.

This synergy stems from the Stage~2 refinement process, which orchestrates multiple agents to collaboratively edit and optimize the entangled Event and Character graphs.
Rather than operating in silos, the agents provide mutual constraints: the Character agent motivates the Plot, while the Theme agent ensures global direction. For example, without this coordination, the Plot might satisfy causal logic but violate personas (e.g., a cowardly character acting heroically), whereas without the Plot agent, character arcs lack structural opportunities to evolve.

\paragraph{Sensitivity Analysis of Iteration Count $K$.}

Table~\ref{tab:k_sensitivity} reports how the maximum number of refinement iterations affects quality and generation scale (GPT-4.1 as backbone). Diversity (Distinct-2) and coherence (Self-BLEU) both peak at $K{=}3$, while Edit-SR remains above 0.9 for $K \le 3$. This supports using a moderate refinement depth to balance quality and controllability; detailed cost--latency trade-offs are discussed in Section~\ref{sec:cost_analysis}. We therefore use $K{=}3$ as default (typically $S{\approx}28$), and keep $K{=}1$ as a budget-oriented alternative (typically $S{\approx}16$).

\begin{table}[h]
\centering
\small
\renewcommand{\arraystretch}{1.1}
\resizebox{\columnwidth}{!}{
\begin{tabular}{ccccccc}
\toprule
$K$ & $S$ & Edit-SR & Tokens & Time & Distinct-2$\uparrow$ & Self-BLEU$\downarrow$ \\
\midrule
1 & 15.8 & 0.96 & 135k & 6.4m  & 0.670 & 0.025 \\
2 & 20.2 & 0.92 & 312k & 10.3m  & 0.710 & 0.020 \\
\textbf{3} & \textbf{27.6} & \textbf{0.91} & \textbf{523k} & \textbf{13.6m}  & \textbf{0.793} & \textbf{0.017} \\
4 & 33.8 & 0.87 & 913k & 21.6m & 0.730 & 0.024 \\
5 & 38.4 & 0.83 & 1257k & 26.1m & 0.640 & 0.035 \\
\bottomrule
\end{tabular}}
\caption{Sensitivity analysis of iteration count $K$. $S$: mean scenes generated. Edit-SR: issue-level success rate that passes the post-edit constraint validation. Bold: default settings.}
\label{tab:k_sensitivity}
\end{table}

\subsection{Computational Cost Analysis}
\label{sec:cost_analysis}
Given that our method involves multiple LLM agents and an iterative cycle, we provide a per-script computational cost breakdown to verify the feasibility of practical deployment in Table~\ref{tab:cost_breakdown}. 
At the default setting ($K{=}3$, GPT-4.1), PLOTTER costs 1.68 USD per script (523k tokens, 13.6 min), suggesting a controllable computational burden while providing superior narrative control.
Notably, Stage~2 (Iterative Refinement) accounts for only 32.4\% of the total cost. 
For budget-constrained scenarios, setting $K{=}1$ further reduces cost to 0.36 USD per script (135k tokens, 6.4 min).

\begin{table}[t]
\centering
\small
\renewcommand{\arraystretch}{1.15}
\begin{tabular}{lc}
\toprule
\textbf{Metric} & \textbf{Value} \\
\midrule
Total cost per script & 1.68 USD \\
Total tokens per script & 523k \\
API calls per script & 171 \\
End-to-end time & 13.6 min \\
\midrule
Stage 1 (Graph Planning) & 4.9\% \\
Stage 2 (Iterative Graph Refinement) & 32.4\% \\
Stage 3 (Script Synthesis) & 62.7\% \\
\midrule
Budget mode ($K{=}1$) cost & 0.36 USD \\
Budget mode time & 6.4 min \\
\bottomrule
\end{tabular}
\caption{Computational cost breakdown.}
\label{tab:cost_breakdown}
\end{table}

\section{Conclusion}
\label{sec:conclusion}
This paper presents \textbf{PLOTTER}, a framework that enhances complex narrative generation by planning on graph-based representations. 
PLOTTER optimizes the causality
and narrative skeleton by diagnosing and repairing structural issues of the graph topology under rigorous logical constraints.
Experiments across diverse scenarios demonstrate that PLOTTER significantly outperforms strong baselines on both storyline logic and full-script quality. 
Ablation studies also reveal a synergy effect, indicating that the co-optimization of plot progression and character psychology yields coherence gains that exceed the sum of isolated improvements. 
These findings show that planning narratives on structural graph representations rather than directly on text is crucial to enhance the long context reasoning in complex narrative generation.

\section*{Limitations}

Although our PLOTTER demonstrates significant improvements in complex narrative generation, there are limitations for future exploration. 
First, the diversity and scale of the evaluation could be further expanded. 
Second, the iterations of the \emph{Evaluate-Plan-Revise} cycle could be further optimized for inference efficiency.  

\section*{Acknowledgments}
This work was partially supported by the National Natural Science Foundation of China under Grants 62306319 and sponsored by CCF-Tencent Rhino-Bird Open Research Fund.

\bibliography{custom}

\appendix

\section{Appendix}
\label{sec:appendix}

\subsection{Definition of Structured Issues}
\label{app:issue_structure}

The analysis function of each agent returns a structured list of issues. To facilitate the graph modification process, each issue is strictly defined with five components, as shown in Table \ref{tab:issue_structure}.

\begin{table}[h] 
    \centering
    \small
    \begin{tabularx}{\linewidth}{l X}
        \toprule
        \textbf{Component} & \textbf{Description} \\
        \midrule
        \textit{Type} & The specific problem category (refer to Table \ref{tab:critic} for the full taxonomy). \\
        \textit{Description} & Detailed identification and explanation of the issue. \\
        \textit{Suggestion} & Concrete optimization guidance provided by the agent. \\
        \textit{Modification} & Required graph operations (e.g., add node, update relation). \\
        \textit{Targets} & The specific graph elements (Nodes/Relations) involved. \\
        \bottomrule
    \end{tabularx}
    \caption{Structure of the issues generated by the Narrative Critics.}
    \label{tab:issue_structure}
\end{table}

\subsection{Dataset Details}
\label{sec:dataset_details}

We sample 50 premises from the following five sources:

\begin{itemize}
    \item \textbf{MoPS}\cite{ma-etal-2024-mops}: A curated collection of movie plot summaries and narrative premises covering diverse genres.
    \item \textbf{WritingPrompts}\cite{fan-etal-2018-hierarchical}: A dataset of user-submitted creative writing prompts designed to stimulate imaginative narrative composition.
    \item \textbf{ROCStories}\cite{mostafazadeh-etal-2016-corpus}: A corpus of commonsense five-sentence stories that follow causal and temporal logic.
    \item \textbf{DOC}: A selection of narrative premises provided in the original DOC paper \cite{yang2023doc}, intended for planning-based narrative generation.
    \item \textbf{LLM generated}: A set of creative premises synthesized using GPT-4.1 across diverse genres and thematic settings.
\end{itemize}

\subsection{Baseline Descriptions}
\label{appendix:baselines}
We compare our method against 3 representative open-source state-of-the-art baselines in script generation:

\begin{itemize}
    \item \textbf{LLM-Plan-Write}: The premise is directly input into the LLM to produce a complete script through narrative planning and writing.

    \item \textbf{Dramatron}: A hierarchical generation method that sequentially generates beats, scenes, and detailed content such as dialogue.

    \item \textbf{DOC}: A planning-based framework that first creates a structured outline and then generates scripts for each part of the outline.
\end{itemize}

\subsection{Evaluation Details}
\label{sec:evaluation_details}

For each pair of scripts generated from the same premise by two different methods, we conduct pairwise comparisons at two levels: (1) by comparing the summaries of corresponding narratives (see Prompt~\ref{prompt:evaluation_storyline}), and (2) by evaluating the overall quality of the full scripts (see Prompt~\ref{prompt:evaluation_full_script}).

\paragraph{Counterbalancing Design.} To ensure that the presentation order does not bias the LLM evaluator's judgment, we randomly split each comparison pair into two groups: one group presents scripts in the order A-B, while the other group presents them in the reverse order B-A, where A and B represent the two methods being compared (i.e., \textbf{PLOTTER} versus one of the baseline methods: LLM-Plan-Write, Dramatron, or DOC). This counterbalancing design ensures that any potential position bias is evenly distributed across both methods, thereby eliminating order effects on the evaluation results.

\paragraph{Evaluation Dimensions.} The evaluator assesses each comparison along five dimensions: (1) \textbf{Narrative}---plot continuity, logical consistency, dramatic arc; (2) \textbf{Thematic Expression}---theme clarity, depth, symbolic reinforcement; (3) \textbf{Characterization}---motivation credibility, psychological depth, character growth; (4) \textbf{Dramatic Engagement}---suspense, turning points, tension management; and (5) \textbf{Premise Fidelity}---adherence to original premise.

\subsection{Human Evaluation Statistics}
\label{sec:appendix_human}
We recruited five professional screenwriters to conduct a human evaluation, aiming to assess the reliability of the LLM evaluation. The final human verdict was determined by majority voting. Table~\ref{tab:human_stats} presents the inter-rater agreement among human evaluators and the alignment statistics between the machine evaluator and human judgments.

\begin{table}[h]
\centering
\small
\renewcommand{\arraystretch}{1.2}
\begin{tabular}{lc}
\toprule
\textbf{Metric} & \textbf{Value} \\
\midrule
\multicolumn{2}{l}{\textit{Inter-Rater Agreement (Human-Human)}} \\
Fleiss' Kappa & 0.688 \\
\midrule
\multicolumn{2}{l}{\textit{Machine-Human Alignment}} \\
Overall Agreement & 90.2\% \\
Cohen's Kappa & 0.834 \\
\bottomrule
\end{tabular}
\caption{Statistics of human evaluation and machine-human consistency.}
\label{tab:human_stats}
\end{table}

The high overall agreement (90.2\%) and Cohen's Kappa (0.834) confirm that the automated metrics used in our main experiments serve as a reliable proxy for human judgment.

\subsection{Case study: End-to-End Graph Evolution}
\label{sec:toy_example}

To concretely demonstrate how the \textsc{Evaluate--Plan--Revise} cycle transforms a narrative, we present a localized walkthrough snapshot generated by PLOTTER ($K{=}1$), highlighting representative edits rather than the full intermediate process.

\begin{figure*}[t]
    \centering
    \includegraphics[width=0.92\linewidth]{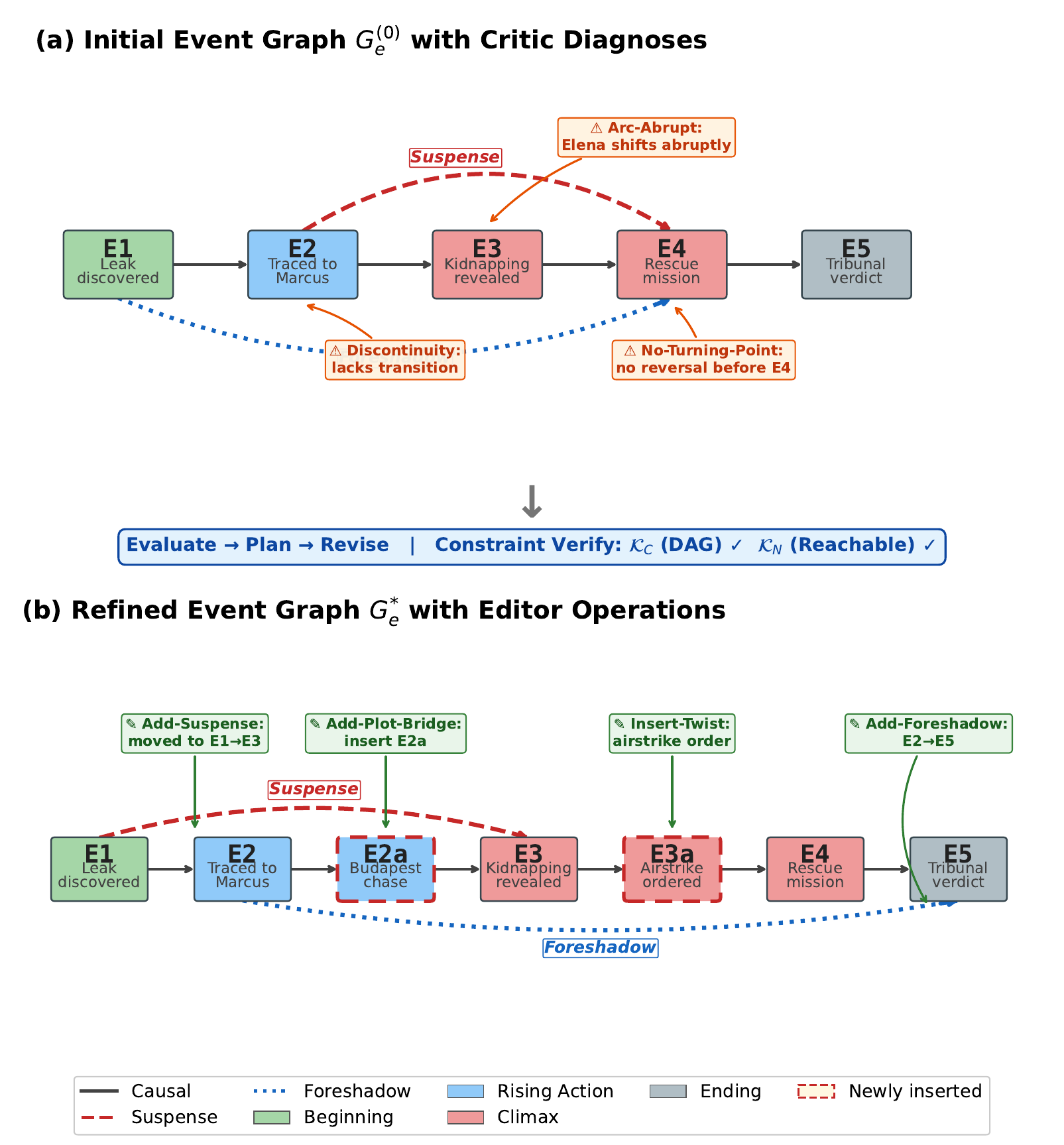}
    \caption{\textbf{Event Graph Evolution (simplified view).} (a) Initial graph $G_e^{(0)}$ with Critic diagnoses: \texttt{Discontinuity} (E2$\to$E3 lacks transition), \texttt{Arc-Abrupt} (Elena shifts abruptly), \texttt{No-Turning-Point} (no reversal before E4). (b) Refined graph $G_e^*$ with Editor operations: \texttt{Add-Plot-Bridge} (E2a inserted), \texttt{Insert-Twist} (E3a airstrike), \texttt{Add-Suspense} (E1$\to$E3), \texttt{Add-Foreshadow} (E2$\to$E5). Dashed red borders indicate newly inserted nodes. The green box confirms both constraints ($\mathcal{K}_C$ DAG, $\mathcal{K}_N$ Reachable) are satisfied. See Table~\ref{tab:toy_events} for the full 16-event inventory.}
    \label{fig:event_graph}
\end{figure*}

Figure~\ref{fig:event_graph} provides a simplified local snapshot of Event Graph evolution, showing how the Critic identifies structural issues (orange callouts) and the Editor resolves them via atomic operations (green callouts). The figure is intentionally partial and condensed; the complete event inventory is provided in Table~\ref{tab:toy_events}.

\subsection{Cross-Genre Generalization}
\label{sec:appendix_genre}

To verify robustness across narrative genres, we analyzed performance stability over the 9 genres in our dataset. The average coefficient of variation (CV) across Distinct-2, MATTR, and Self-BLEU is 0.058, indicating highly stable performance. Furthermore, Suspense edges appear in 96\% of generated scripts and Foreshadowing edges in 100\%, confirming that the graph schema generalizes functionally across genres. Adapting the framework to alternative cultural narrative logics is left for future work.

\begin{table*}[t]
\centering
\small
\renewcommand{\arraystretch}{1.2}
\begin{tabular}{p{2cm}p{3cm}p{8cm}}
\toprule
\textbf{Dimension} & \textbf{Issue Type} & \textbf{Description} \\
\midrule
\multirow{2}{*}{Theme} 
  & \texttt{Theme-Direct} & Theme conveyed only via exposition; lacks symbolic or conflict-driven delivery\\
  & \texttt{Theme-Vague}  & No clear central storyline; events fail to revolve around a core message\\
\midrule
\multirow{3}{*}{Character} 
  & \texttt{Arc-Abrupt}   & Sudden attitude shift without psychological build-up\\
  & \texttt{Motive-Weak}  & Key decisions missing internal/external motivation\\
  & \texttt{One-Dimensional} & Characters show no conflicting traits or growth\\
\midrule
\multirow{5}{*}{Plot} 
  & \texttt{Discontinuity} & Adjacent events lack causal linkage or smooth transition\\
  & \texttt{No-Suspense}  & All information revealed too early; no unanswered questions\\
  & \texttt{No-Foreshadow} & Later twists lack early symbolic hints or setups\\
  & \texttt{No-Turning-Point} & Monotone storyline without rhythm-breaking reversals\\
  & \texttt{Relation-Inconsistent} & Character relations or event logic contradict earlier setup\\
\bottomrule
\end{tabular}
\caption{Issue types identified by \textsc{Multi-Agent Critique}. Each issue is anchored to specific node(s) or edge(s) where the problem occurs.}
\label{tab:critic}
\end{table*}

\begin{table*}[t]
\centering
\small
\renewcommand{\arraystretch}{1.2}
\begin{tabular}{p{3cm}p{3cm}p{7cm}}
\toprule
\textbf{Operation} & \textbf{Action} & \textbf{Purpose} \\
\midrule
\texttt{Add-Plot-Bridge} & Insert intermediate event node & Repair logical gaps between consecutive events; address discontinuities \\
\texttt{Add-Suspense}  & Insert mystery/misleading clue & Enhance narrative tension; address low tension and lack of suspense \\
\texttt{Add-Foreshadow} & Embed symbolic detail or hint & Prepare for later narrative payoff; resolve lack of foreshadowing \\
\texttt{Insert-Twist} & Add unexpected but logical reversal & Introduce narrative rhythm change; address monotony \\
\texttt{Revise-Event} & Modify existing node/relation & Harmonize motivation chains; strengthen thematic consistency\\
\bottomrule
\end{tabular}
\caption{Atomic edit operations executed by \textsc{Constrained Graph Editor}. Edit plans may combine multiple operations while satisfying scope and causality constraints.}
\label{tab:editor}
\end{table*}

\begin{table*}[t]
\centering
\small
\renewcommand{\arraystretch}{1.12}
\setlength{\tabcolsep}{3.5pt}
\begin{tabular}{cl c p{8.2cm} p{3.8cm}}
\toprule
\textbf{ID} & \textbf{Stage} & \textbf{Time} & \textbf{Description} & \textbf{Edit Operation} \\
\midrule
\multicolumn{5}{l}{\textit{Original Events (from $G_e^{(0)}$)}} \\
\midrule
A3K & Beginning      & Day 1  & Elena and Marcus complete a Prague extraction; Elena intercepts an encrypted leak from their safe house to enemy syndicate Aegis. & --- \\
F8W & Rising Action  & Day 3  & Analyst Nadia traces the leak to Marcus's device; a two-year pattern of compromised missions emerges. & --- \\
P2R & Rising Action  & Day 5  & Vienna bait-trap confirms Marcus as the mole. He vanishes, leaving a note: ``They have her. Forgive me.'' & --- \\
J6T & Climax         & Day 8  & Budapest confrontation at gunpoint. Marcus reveals Aegis kidnapped his daughter Lily two years ago. Elena hesitates; he escapes. & \textit{Modified}: deepened Elena's internal conflict \\
M4V & Falling Action & Day 10 & Elena files an incomplete report omitting Lily. Graves assigns Kessler to shadow her; Nadia warns Graves may have known. & --- \\
H9B & Rising Action  & Day 13 & Via dead-drops, Marcus reveals Lily's Carpathian location. Elena agrees to help if he surrenders afterward. & --- \\
Q1X & Rising Action  & Day 15 & Graves orders an airstrike despite Lily being inside. Elena strikes Kessler and defects from headquarters. & \textit{Modified}: added Elena's one-word refusal \\
D5Z & Climax         & Day 16 & Night assault on the compound. Elena kills Viktor; Marcus reunites with Lily; they escape before the airstrike. & --- \\
W7G & Falling Action & Day 19 & Elena exposes Graves's secret Aegis dealings using gathered intelligence. Graves is arrested for treason. & --- \\
E2N & Ending         & Day 25 & Disgraced Elena testifies at Marcus's tribunal. Reduced sentence; she is dismissed but promises to watch over Lily. & \textit{Modified}: expanded with legal consequences \\
\midrule
\multicolumn{5}{l}{\cellcolor{green!8}\textit{Newly Inserted Events (by Stage~2 Constrained Graph Editor)}} \\
\midrule
\rowcolor{green!8}
K7P & Falling Action & Day 11 & Elena breaks into archives and finds Graves's memo labeling Lily ``acceptable operational attrition'' and his override of Marcus's psych evaluation. & \texttt{Add-Plot-Bridge}: \textit{Discontinuity} (M4V$\to$H9B) \\
\rowcolor{green!8}
T3M & Rising Action  & Day 14 & Night before rescue. Marcus shows Lily's crayon drawings and birthday video. Elena's institutional loyalty dissolves. & \texttt{Add-Plot-Bridge}: \textit{Arc-Abrupt} (Elena) \\
\rowcolor{green!8}
U2F & Rising Action  & Day 15\textsuperscript{am} & Elena returns to HQ for satellite data; Graves confronts her with intercepted dead-drop transcripts. Her cover collapses. & \texttt{Add-Plot-Bridge}: \textit{Discontinuity} (H9B$\to$Q1X) \\
\rowcolor{green!8}
V4Q & Rising Action  & Day 15\textsuperscript{pm} & Minutes after fleeing HQ, Elena's hands shake. Lily's photo on her phone sustains her---she drives forward despite terror. & \texttt{Revise-Event}: \textit{One-Dimensional} (Elena) \\
\rowcolor{green!8}
R3Y & Rising Action  & Day 15\textsuperscript{night} & En route, Elena passes through ruins of an agency-destroyed village. A child's shoe on a threshold crystallizes her resolve. & \texttt{Add-Foreshadow}: \textit{No-Foreshadow} (D5Z) \\
\rowcolor{green!8}
X3L & Falling Action & Day 21 & Prosecutor offers a deal: blame Marcus, get reinstated. Elena refuses, choosing to testify as a disgraced agent with no leverage. & \texttt{Add-Plot-Bridge}: \textit{Motive-Weak} (W7G$\to$E2N) \\
\bottomrule
\end{tabular}
\caption{\textbf{Event Graph: Initial $\to$ Refined.} Top: 10 original events from $G_e^{(0)}$. Bottom (\colorbox{green!8}{shaded}): 6 events inserted by the Editor during Stage~2, each annotated with the Critic issue type that triggered the insertion and the specific gap it bridges.}
\label{tab:toy_events}
\end{table*}

\begin{figure*}[t]
    \centering
    \includegraphics[width=\linewidth]{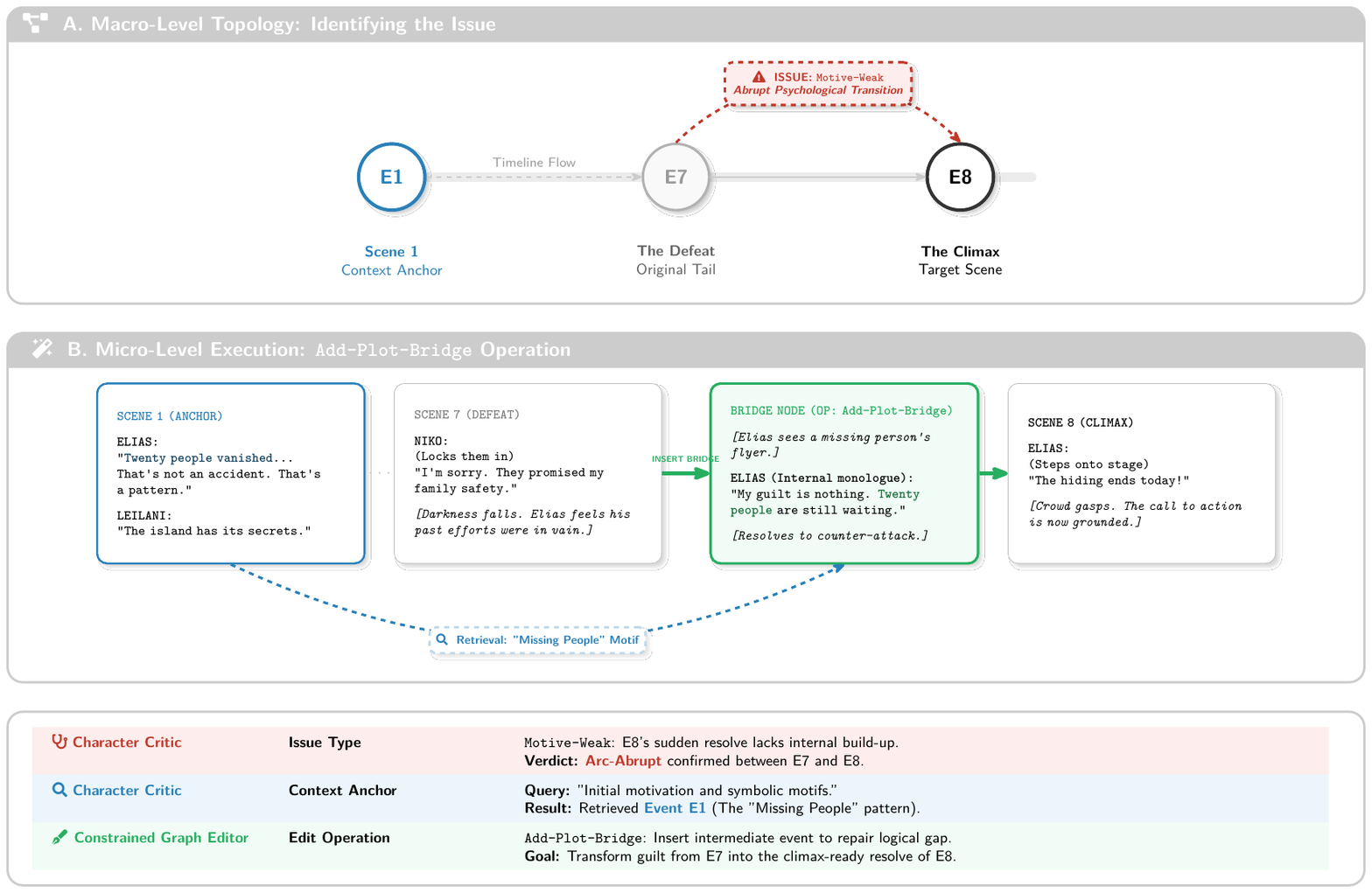}
    \caption{\textbf{Context-Aware Diagnosis and Retrieval.} 
    (A) \textbf{Macro-Level Topology}: The \textsc{Multi Agent Critic} identifies a logical break between the defeat in E7 and the confidence in E8, flagging a \texttt{Motivation Gap} (see Issue Types in Table~\ref{tab:critic}).
    (B) \textbf{Micro-Level Execution}: The \textsc{Constrained Graph Editor} performs a historical query to retrieve the "Missing People" motif, executing a context-aware \texttt{Add-Plot-Bridge} operation (Table~\ref{tab:editor}).
    This allows the system to generate a specific bridging scene where Elias converts fear into anger.}
    \label{fig:case_topology}
\end{figure*}

\begin{figure*}[t]
    \centering
    \includegraphics[width=\linewidth]{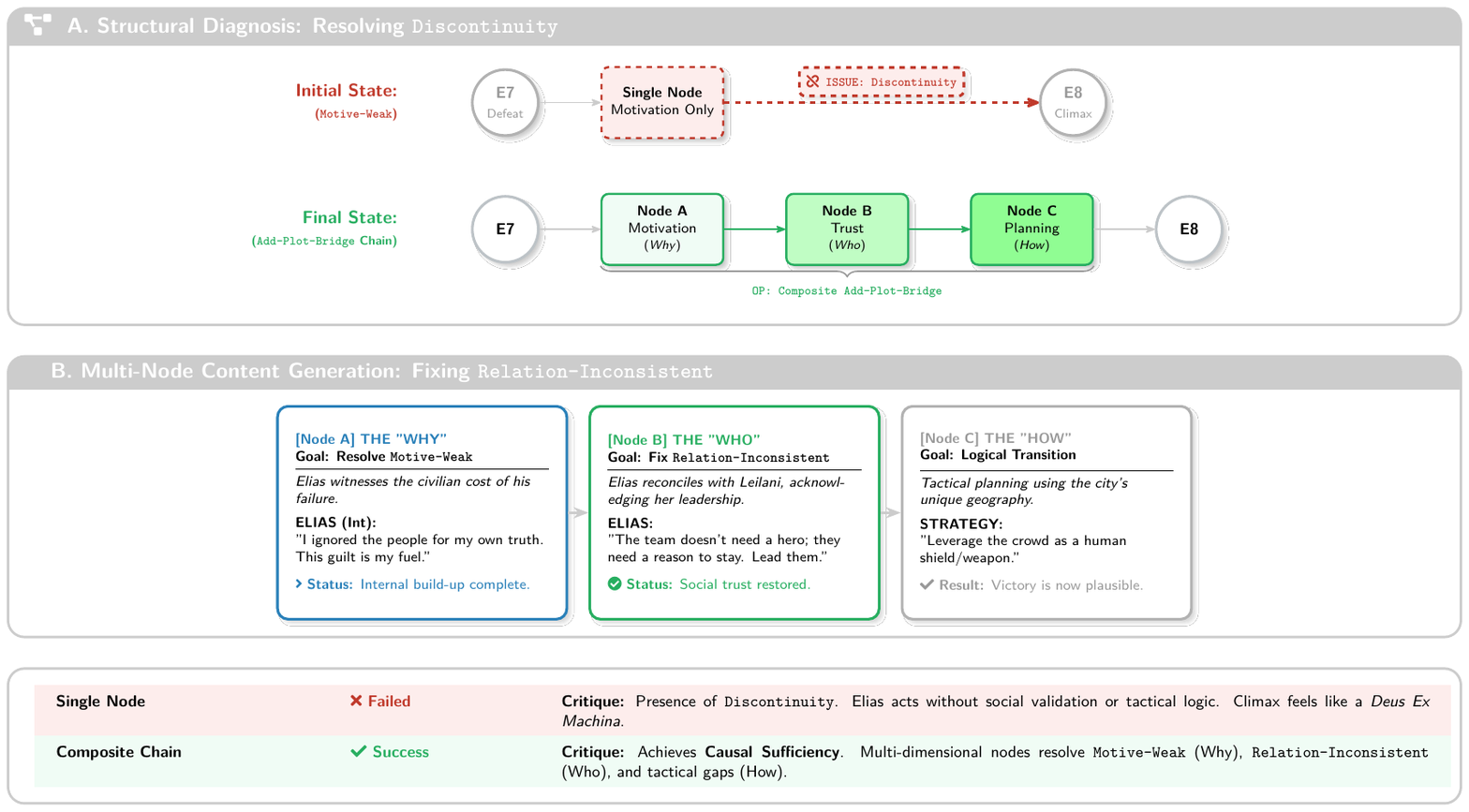}
    \caption{\textbf{From Single Node to Causal Chain.} 
    (A) \textbf{Structural Diagnosis}: A simple single-node insertion (Ins-A) fails because it only addresses internal feelings.
    (B) \textbf{The "Trinity" of Action}: The Full Module generates a multi-hop causal chain by iteratively applying \texttt{Add-Plot-Bridge} operations: \textit{Motivation} (Ins-A) $\to$ \textit{Trust} (Ins-B) $\to$ \textit{Planning} (Ins-C).
    This ensures the victory in E8 is logically earned.}
    \label{fig:case_structural}
\end{figure*}

\begin{table*}[t]
\centering
\small
\setlength{\tabcolsep}{8pt}
\begin{tabular}{p{3.2cm}p{1.3cm}p{10cm}}
\toprule
\textbf{Dimension} & \textbf{Verdict} & \textbf{Evaluator Explanation} \\
\midrule
Narrative & \cellcolor{blue!10}\textbf{B} & Script B demonstrates a more intricate and logically consistent narrative structure, with clear exposition, rising action, multiple points of conflict, and resolution, as well as smoother transitions and deeper world-building, whereas Script A, while coherent, follows a more straightforward and less nuanced dramatic arc. \\
\addlinespace[3pt]
Thematic Expression & \cellcolor{blue!10}\textbf{B} & Script B demonstrates a deeper and more nuanced exploration of its central theme of redemption and trust, employing recurring metaphors (such as the weathered whistle, anonymous notes, and symbolic gestures like offering the cap or playbook) and moments of self-reflection to reinforce the theme, while Script A, though consistent, approaches its theme more directly and with less subtlety or symbolic layering. \\
\addlinespace[3pt]
Characterization & \cellcolor{blue!10}\textbf{B} & Script B demonstrates greater emotional and psychological depth by exploring Cameron's internal struggles, the impact of his past on his present relationships, and the nuanced, gradual development of trust between him and Elijah, resulting in a more complex and believable character arc with clear growth for both father and son. \\
\addlinespace[3pt]
Dramatic Engagement & \cellcolor{blue!10}\textbf{B} & Script B demonstrates a higher level of dramatic engagement by introducing suspenseful elements such as anonymous notes and rumors, multiple significant turning points, and a more intricate buildup and release of tension through both personal and team-related stakes, whereas Script A, while emotionally resonant, follows a more straightforward and predictable arc with fewer suspenseful twists. \\
\addlinespace[3pt]
Premise Fidelity & \cellcolor{gray!10}\textbf{Same} & Both scripts faithfully adhere to the core premise, retaining the primary setting (high school football), central characters (Cameron, Elijah, and the mother), and the thematic direction of a former NFL player grappling with fatherhood and coaching; neither script significantly deviates from or adds extraneous elements that undermine the stated premise. \\
\bottomrule
\end{tabular}
\caption{GPT-4.1 pairwise evaluation results for a representative case. Premise: \textit{``A former NFL player grapples with fatherhood and coaching his son in high school football.''} Script A corresponds to the baseline (Dramatron), while Script B corresponds to \textbf{PLOTTER}. Blue cells indicate Script B superiority; gray indicates equal performance.}
\label{tab:case_comparison}
\end{table*}

\begin{table*}[t]
\centering
\renewcommand{\arraystretch}{1.25}
\scriptsize
\setlength{\tabcolsep}{5pt}
\begin{tabular}{p{0.12\textwidth}|p{0.38\textwidth}|p{0.38\textwidth}}
\toprule
\textbf{Narrative Function} & \textbf{Script A: "Second Chances"} & \textbf{Script B: "Second Down"} \\
\midrule
\rowcolor{gray!5}
\textbf{Opening \& Discovery} & 
\textbf{S1:} Cameron notices a quiet boy watching practice alone and learns, to his shock, that the boy is Elijah—his son. & 
\textbf{S1:} Cameron, newly retired from the NFL, returns as coach to his old high school. As he fingers a \textbf{weathered whistle} by the trophy case, he confronts his regret and hopes for redemption. \newline
\textbf{S2:} After practice, Cameron discovers an \textbf{anonymous note} in his locker referencing a past mistake and warning him that 'the past always catches up.' \newline
\textbf{S3:} Cameron notices Elijah watching practice from afar. Feeling a strange familiarity, he considers inviting Elijah closer with a team \textbf{cap}, but hesitates. \\
\midrule
\textbf{Confrontation \& Truth Revealed} & 
\textbf{S2:} Cameron confronts Elijah's mother, Jenna, demanding answers about why he never knew about his son, sparking anger and resentment. \newline
\textbf{S3:} Cameron debates whether to pursue a relationship with Elijah or respect Jenna's wishes to keep things as they are. & 
\textbf{S4:} Lena confronts Cameron, revealing that Elijah is his son. Cameron reels from the shock, overwhelmed as Lena insists on honesty. (The \textbf{cap} drops from his hand) \\
\midrule
\rowcolor{gray!5}
\textbf{External Pressure \& Crisis} & 
\textbf{S4:} The principal questions Cameron's commitment to coaching after learning about his personal distractions. \newline
\textbf{S5:} Cameron's struggle to focus causes him to snap at a struggling player, damaging team morale. & 
\textbf{S5:} Cameron's attempt to connect with Elijah is rebuffed. The weight of the \textbf{anonymous note} and his guilt pushes Cameron to a breaking point. \newline
\textbf{S6:} The aftermath of Cameron's failures ripple through the team. Lena urges Cameron to truly commit to Elijah if he wants a real relationship. \\
\midrule
\textbf{Internal Struggle \& Reflection} & 
\textbf{S6:} Cameron tries to connect with Elijah over breakfast, but Elijah remains distant and untrusting. \newline
\textbf{S7:} Jenna accuses Cameron of trying to make up for the past too quickly, fueling guilt and self-doubt. & 
\textbf{S7:} Haunted by Lena's words, Cameron reflects on how easily trust can be broken, realizing that real growth demands vulnerability. \newline
\textbf{S8:} Moved by self-reflection, Cameron gently reaches out to Elijah, sharing personal stories from his troubled youth. \\
\midrule
\rowcolor{gray!5}
\textbf{Complications \& Setbacks} & 
\textbf{S8:} Cameron learns that Elijah is being bullied for being the coach's 'secret' kid, escalating Cameron's inner conflict. \newline
\textbf{S9:} Cameron witnesses Elijah stand up for himself, revealing resilience but also anger toward Cameron. \newline
\textbf{S10:} Cameron is forced to choose between supporting his struggling team and comforting Elijah, ultimately failing both. & 
\textbf{S9:} After reconnecting with Elijah, Cameron overhears players discussing a possible sabotage for the playoff game. (Cameron's NFL \textbf{cap} is mentioned in context) \newline
\textbf{S10:} Elijah, conflicted by Cameron's efforts, confides in Lena about his doubts and hope. He cautiously decides to give Cameron a chance. \\
\midrule
\textbf{Tentative Reconciliation} & 
\textbf{S11:} Elijah tentatively joins practice, and Cameron struggles to maintain professional boundaries while connecting as a father. & 
\textbf{S11:} Elijah takes a tentative step by joining the team as a waterboy. Cameron struggles to balance his roles. (Scene contains both \textbf{whistle} and \textbf{cap}) \\
\midrule
\rowcolor{gray!5}
\textbf{Pre-Climax Preparation} & 
\textbf{S12:} A key game arrives, and Cameron must decide between prioritizing victory and supporting Elijah emotionally. \newline
\textbf{S13:} During halftime, Cameron delivers a motivational speech that rallies the team but leaves Elijah feeling sidelined. & 
\textbf{S12:} Cameron, tested by Elijah's hesitation and the team's skepticism, commits to vulnerable leadership and transparency. \newline
\textbf{S13:} Before the critical game, Cameron and Elijah share a quiet moment. Cameron's gesture of solidarity begins to rebuild trust. \\
\midrule
\textbf{Climax \& Key Decision} & 
\textbf{S14:} After the game, Cameron apologizes publicly to Elijah, acknowledging his failures as a father. \newline
\textbf{S15:} Cameron and Elijah reconcile as the team wins. Cameron realizes that success is measured by relationships, not trophies. & 
\textbf{S14:} During the game, the team works together despite challenges. Cameron's consistent support of Elijah shifts the team dynamic. \newline
\textbf{S15:} When a key player is injured, Elijah steps up to help. Cameron faces a pivotal decision: trust Elijah with the \textbf{playbook} and his coaching strategy, symbolizing complete vulnerability and trust. \\
\midrule
\rowcolor{blue!8}
\textbf{Extended Resolution} & 
\textit{[Script A ends at S15]} & 
\textbf{S16:} The team rallies around Elijah's contribution. Cameron and Elijah's relationship deepens through shared responsibility. \newline
\textbf{S17:} Post-game, Cameron and Elijah reflect on their journey. Cameron acknowledges his past while embracing his role as father. \newline
\textbf{S18:} The team celebrates their hard-fought victory. Cameron and Lena discuss co-parenting, establishing healthier boundaries. \newline
\textbf{S19:} In a quiet moment on the field, Cameron and Elijah toss the football. The \textbf{whistle}, \textbf{cap}, and \textbf{playbook} are all visible, symbolizing their completed journey from strangers to family. \\
\bottomrule
\end{tabular}
\caption{Narrative-function-aligned comparison for the representative case. Scenes are grouped by their dramatic role (e.g., opening, confrontation, climax). Script A uses 15 scenes total; Script B uses 19 scenes with finer-grained division within each narrative function. Recurring symbolic elements in Script B are highlighted in \textbf{bold}. The blue row shows Script B's extended resolution (4 additional scenes).}
\label{tab:script_excerpts}
\end{table*}

\onecolumn
\begin{small}
\begin{longtable}{p{0.95\textwidth}}
\toprule
\textbf{Script Title Generation Prompt:} \\
\texttt{Script Summary: \{storyline\}}\\
\texttt{Generate a \textbf{concise and distinctive} title for this script (no more than 7 words).}\\
\texttt{The title must reflect the \textbf{theme, genre, and narrative structure} of the script.}\\
\texttt{It should be \textbf{expressive, poetic, and meaningful}, in the spirit of classic film titles.}\\
\texttt{Additionally, the title should be \textbf{captivating and marketable}, suitable for \textbf{film posters or streaming platforms}.}\\
\vspace{0.1cm}
\texttt{\textbf{Example Titles:}}\\
\texttt{1. Summary: A war correspondent exposes a military cover-up and is hunted as a traitor by the government.}\\
\texttt{   Generated Title: "Shadow in the Fire"}\\
\texttt{2. Summary: A scientist discovers time travel but realizes using it will erase his daughter from history.}\\
\texttt{   Generated Title: "The Vanishing Hour"}\\
\texttt{3. Summary: In a dystopian city where dreaming is forbidden, a woman risks everything to protect the last dreamer.}\\
\vspace{0.1cm}
\texttt{   Generated Title: "The Last Dream"}\\
\texttt{Script Summary: \{storyline\}}\\
\texttt{Return strictly in JSON format:}\\
\texttt{\{"title": "Generated Title"\}}\\
\bottomrule
\caption{Prompt for generating a concise and distinctive title for a script.}
\label{prompt:title}
\end{longtable}
\end{small}

\begin{small}
\begin{longtable}{p{0.95\textwidth}}
\toprule
\textbf{Plot and Character Graph Creation Prompt:} \\
\texttt{script Title: \{title\}}\\
\texttt{Premise: \{premise\}}\\
\texttt{Based on this premise, generate a Event Graph and a Character Graph.}\\
\vspace{0.1cm}
\texttt{\textbf{Event Graph:}}\\
\texttt{  - The plot must have a clear main storyline with a complete arc from the 'Beginning' to the 'Ending'.}\\
\texttt{  - Each event node must have a clear causal relationship with the next, forming a tightly connected chain.}\\
\texttt{  - The plot should contain around 10 events to maintain narrative focus and pacing.}\\
\texttt{  - There must be two Climax nodes (with \textbackslash{}texttt\{narrative\_stage = "Climax"\}) positioned at two major narrative peaks.}\\
\texttt{  - The event structure must strictly follow the sequence and count below (10 nodes in total):}\\
\texttt{    1. Beginning (1)}\\
\texttt{    2. First Rising Actions (2–3)}\\
\texttt{    3. First Climax (4)}\\
\texttt{    4. First Falling Action (5)}\\
\texttt{    5. Second Rising Actions (6–7)}\\
\texttt{    6. Second Climax (8)}\\
\texttt{    7. Second Falling Action (9)}\\
\texttt{    8. Ending (10)}\\
\texttt{  - The 'Beginning' must be the first event node, and the 'Ending' must be the last.}\\
\texttt{  - Each event must include the following narrative attributes:}\\
\texttt{    - \{narrative\_stage\}: One of the following:}\\
\texttt{      - Beginning: Introduces time, setting, and main characters; shows initial status quo and a triggering problem. Only the first node.}\\
\texttt{      - Rising Action: Characters try to solve the problem but face obstacles; tension and stakes increase.}\\
\texttt{      - Climax: The most intense point of conflict or a major turning point. Must appear twice.}\\
\texttt{      - Falling Action: Consequences unfold; tensions ease; relationships and power dynamics shift.}\\
\texttt{      - Ending: Final state of the protagonist; answers the initial question and reflects growth. Only the last node.}\\
\texttt{  - Each event must include:}\\
\texttt{    - \{id\}: Unique identifier for the event}\\
\texttt{    - \{description\}: A full description of the event, including character actions}\\
\texttt{    - \{narrative\_stage\}: One of the five stages defined above}\\
\texttt{    - \{time\}: Time the event occurs (e.g., "Day 1")}\\
\vspace{0.1cm}
\texttt{\textbf{Character Graph:}}\\
\texttt{- Characters must have rich, multi-dimensional relationships, not just 'friend' or 'enemy'. Include:}\\
\texttt{  - Conflict Relations (rivalry, revenge, misunderstanding)}\\
\texttt{  - Cooperative Relations (ally, mentor-mentee, colleague)}\\
\texttt{  - Emotional Relations (family, romantic, unspoken affection)}\\
\texttt{  - Hidden Relations (secret identity, double agent, concealed hostility)}\\
\texttt{- Classify characters into:}\\
\texttt{  - Main Characters: Drive the plot and make key decisions that shape the outcome.}\\
\texttt{  - Supporting Characters: Expand the world and support the narrative through loyalty shifts, clue delivery, or creating conflict.}\\
\texttt{- All relationships should have logical narrative motivation—avoid random or unexplained links.}\\
\texttt{- The Character Graph must align with the Event Graph and ensure character presence is justified within events.}\\
\vspace{0.1cm}
\texttt{\textbf{Example Output in JSON (Strictly follow this format):}}\\
\texttt{\{}\\
\texttt{  "plot\_graph": \{}\\
\texttt{    "events": [}\\
\texttt{      \{"id": "E1", "description": "...", "narrative\_stage": "Beginning", "time": "Day 1"\},}\\
\texttt{      \{"id": "E2", "description": "...", "narrative\_stage": "Rising Action", "time": "Day 2"\},}\\
\texttt{      ...}\\
\texttt{    ],}\\
\texttt{    "relations": [}\\
\texttt{      \{"from": "E1", "to": "E2", "relation": "causal"\},}\\
\texttt{      ...}\\
\texttt{    ]}\\
\texttt{  \},}\\
\texttt{  "character\_graph": \{}\\
\texttt{    "characters": [}\\
\texttt{      \{"id": "C1", "name": "Jack", "motivation": "..." \},}\\
\texttt{      ...}\\
\texttt{    ],}\\
\texttt{    "relations": [}\\
\texttt{      \{"from": "C1", "to": "C2", "relation": "..." \},}\\
\texttt{      ...}\\
\texttt{    ]}\\
\texttt{  \}}\\
\texttt{\}}\\
\bottomrule
\caption{Instructional prompt used to generate the Event Graph and Character Graph structures from a given script title and premise.}
\label{prompt:story_graph}
\end{longtable}
\end{small}
\begin{small}
\begin{longtable}{p{0.95\textwidth}}
\toprule
\textbf{Character Creation Prompt:} \\
\texttt{You are generating characters for a script. Ensure that the characters fit the tone, style, and emotional depth of the script.}\\
\vspace{0.1cm}
\texttt{\textbf{Guidelines for Character Creation:}}\\
\texttt{- \textbf{Each character must have:}}\\
\texttt{1. \textbf{A core personality trait} (e.g., loyal, ambitious, paranoid, cynical).}\\
\texttt{2. \textbf{An internal conflict (fatal flaw)} (e.g., pride, guilt, fear, obsession).}\\
\texttt{3. \textbf{An external goal} (e.g., expose the truth, save a loved one, escape the past).}\\
\texttt{4. \textbf{A relationship with at least one other character} (mentor, enemy, lost love, etc.).}\\
\texttt{5. \textbf{A hidden past or secret} that impacts their choices.}\\
\vspace{0.1cm}
\texttt{\textbf{Example:}}\\
\texttt{Storyline: A scientist discovers time travel but realizes using it will erase his own daughter from existence.}\\
\texttt{Generated Characters:}\\
\texttt{\{}\\
\texttt{  "characters": \{}\\
\texttt{    "Dr. Monroe": \{}\\
\texttt{      "description": "A brilliant physicist (obsessive, morally torn), willing to break time itself to save his daughter. His guilt over a past mistake fuels his desperation."}\\
\texttt{    \},}\\
\texttt{    "Evelyn Monroe": \{}\\
\texttt{      "description": "A rebellious teenager (curious, fearless), unaware that her father’s invention is rewriting her existence."}\\
\texttt{    \},}\\
\texttt{    "Agent Carter": \{}\\
\texttt{      "description": "A government enforcer (ruthless, efficient) who believes time travel is a weapon. He once saved Monroe’s life, but now sees him as a threat."}\\
\texttt{    \}}\\
\texttt{  \}}\\
\texttt{\}}\\
\vspace{0.1cm}
\texttt{Title: \{title\}}\\
\texttt{Storyline: \{storyline\}}\\
\texttt{Character Graph:\{character\_graph\}}\\
\texttt{Strictly return JSON format:}\\
\texttt{\{"characters": \{"Character Name": \{"description": "Short but rich character description"\}\}\}}\\
\bottomrule
\caption{Prompt used for generating characters in a script.}
\label{prompt:character_expansion}
\end{longtable}
\end{small}

\begin{small}
\begin{longtable}{p{0.95\textwidth}}
\toprule
\textbf{Scene Generation Prompt:} \\
\texttt{You are generating a \textbf{structured sequence of scenes} for a script.}\\
\texttt{Each scene must be \textbf{based on the events from the Event Graph} and take into account the \textbf{character relationships from the Character Graph}.}\\
\vspace{0.1cm}
\texttt{\textbf{Script Information:}}\\
\texttt{\textbf{Title}: \{title\}}\\
\texttt{\textbf{Premise}: \{premise\}}\\
\texttt{\textbf{Event Graph}:}\\
\texttt{\{plot\_description\}}\\
\texttt{\textbf{Character Graph}:}\\
\texttt{\{character\_description\}}\\
\vspace{0.1cm}
\texttt{\textbf{Scene Generation Requirements:}}\\
\texttt{- \textbf{Each scene must correspond to an event in the event graph}, following the same order.}\\
\texttt{- Ensure \textbf{clear causal connections between events} and maintain chronological order.}\\
\texttt{- The \textbf{protagonist and key characters must appear in relevant scenes}, consistent with the character graph.}\\
\texttt{- \textbf{Each scene must have a clear objective} and contribute to the advancement of the narrative.}\\
\texttt{- The number of generated scenes should match the number of events in the event graph.}\\
\vspace{0.1cm}
\texttt{\textbf{Example:}}\\
\texttt{\textbf{Premise}: In a futuristic city, a hacker uncovers a massive conspiracy after breaching the government system.}\\
\texttt{Generated scenes:}\\
\texttt{[}\\
\texttt{\{}\\
\texttt{"place": "Neon-lit alley in the cyberpunk city",}\\
\texttt{"plot\_element": "Inciting Incident",}\\
\texttt{"beat": "Hacker Logan discovers an encrypted file revealing the government's dark secrets."}\\
\texttt{\},}\\
\texttt{\{}\\
\texttt{"place": "Underground resistance base",}\\
\texttt{"plot\_element": "Climax",}\\
\texttt{"beat": "Logan must decide whether to release the file, risking his life."}\\
\texttt{\}}\\
\texttt{]}\\
\vspace{0.1cm}
\texttt{Strictly return a JSON list in the following format:}\\
\texttt{[}\\
\texttt{\{"place": "Scene location", "plot\_element": "Narrative function", "beat": "Key moment that drives the plot"\}, ...}\\
\texttt{]}\\
\bottomrule
\caption{Prompt for generating a structured sequence of scenes for a script.}
\label{prompt:scene_batch}
\end{longtable}
\end{small}

\begin{small}
\begin{longtable}{p{0.95\textwidth}}
\toprule
\textbf{Graph-Enhanced Scene Description Generation Prompt:} \\
\texttt{You are generating a \textbf{cinematic and immersive scene description} for a script that integrates narrative graph structure.}\\
\vspace{0.1cm}
\texttt{\textbf{Scene Context:}}\\
\texttt{- \textbf{Location}: \{scene.place\}}\\
\texttt{- \textbf{Plot Element}: \{scene.plot\_element\}}\\
\texttt{- \textbf{Key Moment}: \{scene.beat\}}\\
\texttt{- \textbf{Scene Position in Narrative Arc}: Scene \{scene\_index + 1\}}\\
\vspace{0.1cm}
\texttt{\textbf{Event Graph Context:}}\\
\texttt{\{graph\_context\}}\\
\vspace{0.1cm}
\texttt{\textbf{Scene Description Guidelines:}}\\
\texttt{- \textbf{Describe the setting in a way that enhances the emotional tone} and reflects the narrative position in the graph structure.}\\
\texttt{- \textbf{Incorporate sensory details} that align with the causal and suspense relations from the narrative graph.}\\
\texttt{- \textbf{Show how characters interact with the environment} in ways that reflect their graph-encoded motivations and relationships.}\\
\texttt{- \textbf{Make sure the scene description hints at narrative tension} that respects the graph's suspense and foreshadowing edges.}\\
\texttt{- \textbf{Reference thematic elements} that emerge from the graph structure if appropriate.}\\
\vspace{0.1cm}
\texttt{\textbf{Graph-Informed Guidelines:}}\\
\texttt{- If this scene follows a suspense edge in the graph, build atmospheric tension through environmental description.}\\
\texttt{- If this scene follows a foreshadowing edge, include subtle visual elements that hint at future developments.}\\
\texttt{- If this scene is a causal continuation, ensure the setting logically follows from previous scenes.}\\
\vspace{0.2cm}
\texttt{Strictly return a JSON object:}\\
\texttt{\{"scene\_description": "A concise and vivid scene description that integrates graph structure."\}}\\
\bottomrule
\caption{Prompt for generating graph-enhanced scene descriptions that leverage narrative graph topology.}
\label{prompt:scene_description}
\end{longtable}
\end{small}

\begin{small}
\begin{longtable}{p{0.95\textwidth}}
\toprule
\textbf{Multi-Issue Plot and Character Modification Prompt:} \\
\texttt{Current Event Graph:}\\
\texttt{\{event\_graph\}}\\
\vspace{0.2cm}
\texttt{Current Character Graph:}\\
\texttt{\{character\_graph\}}\\
\vspace{0.2cm}
\texttt{Below is a list of multiple issues that need to be addressed. Please generate a corresponding modification plan for **each individual issue**, based on its **type**, **description**, **suggested solution**, **modification method**, **involved nodes**, and **involved relations**:}\\
\texttt{Issue List:}\\
\texttt{\{parsed\_issues\}}\\
\vspace{0.2cm}
\texttt{Please follow these modification constraints:}\\
\begin{itemize}
  \item \texttt{\textbf{Scope Constraint}: Only modify nodes and relations explicitly listed in "involved nodes" and "involved relations" fields.}
  \item \texttt{\textbf{Method Constraint}: All modifications must follow the "modification method" field; no additional changes are permitted.}
  \item \texttt{\textbf{Node Insertion}: New nodes are only allowed when explicitly permitted by the modification method. New nodes must have unique IDs and maintain narrative stage consistency with adjacent nodes.}
  \item \texttt{\textbf{Relation Types}: Use appropriate relation types (causal, suspense, foreshadowing) based on the narrative function of the modification.}
\end{itemize}\\

\vspace{0.2cm}
\texttt{\#\#\# Example Output Format:}\\
\texttt{\{}\\
\texttt{"issues": [}\\
\texttt{\{}\\
\texttt{"issue\_id": 1,}\\
\texttt{"plot\_changes": \{}\\
\texttt{"Delete relation": [\{"from": "E2", "to": "E3"\}],}\\
\texttt{"New event": [\{"id": "E2a", "description": "...", "narrative\_stage": "Rising Action", "time": "..."\}],}\\
\texttt{"Modify event": [\{"id": "E3", "description": "...", "narrative\_stage": "...", "time": "..."\}],}\\
\texttt{"New relation": [\{"from": "E2", "to": "E2a", "relation": "causal"\}, \{"from": "E2a", "to": "E3", "relation": "causal"\}]}\\
\texttt{\},}\\
\texttt{"character\_changes": \{}\\
\texttt{"Delete relation": [\{"from": "C1", "to": "C2"\}],}\\
\texttt{"New character": [\{"id": "C3", "name": "...", "motivation": "..."\}],}\\
\texttt{"Modify character": [\{"id": "C1", "name": "...", "motivation": "..."\}],}\\
\texttt{"New relation": [\{"from": "C1", "to": "C3", "relation": "..."\}]}\\
\texttt{\}}\\
\texttt{\}}\\
\texttt{]}\\
\texttt{\}}\\
\bottomrule
\caption{Prompt for generating modification plans for plot and character issues.}
\label{prompt:incremental_improvement}
\end{longtable}
\end{small}

\begin{small}
\begin{longtable}{p{0.95\textwidth}}
\toprule
\textbf{Theme Agent Analysis Prompts:} \\
\vspace{0.2cm}
\texttt{The Theme Agent evaluates two dimensions: (1) theme clarity and (2) thematic expression explicitness.}\\
\vspace{0.1cm}
\textbf{1. Theme Missing Analysis:}\\
\texttt{You are a script theme structure analyst. Please evaluate the overall event graph to identify whether there is an issue of \textbf{"Unclear Theme or Storyline"}, and provide a suggestion to strengthen the thematic coherence by modifying the content of existing event nodes.}\\
\vspace{0.1cm}
\texttt{\textbf{Problem Definition:}}\\
\texttt{\textbf{Unclear Theme or Storyline}: The script lacks a clearly defined, consistently developed central theme or narrative thread. As a result, character actions and plot developments do not revolve around a central issue or narrative direction.}\\
\vspace{0.1cm}
\texttt{\textbf{Optimization Guidelines:}}\\
\texttt{- Do not add any new event nodes;}\\
\texttt{- Refine the content of existing nodes to reveal and emphasize an underlying thematic focus;}\\
\texttt{- Ensure that multiple key nodes reflect the same core idea through character dilemmas and decisions;}\\
\texttt{- Output only one issue.}\\
\vspace{0.1cm}
\textbf{2. Theme Explicitness Analysis:}\\
\texttt{You are an expert in optimizing thematic expression in screenwriting. Please analyze the following event graph to determine whether there is an issue of \textbf{"Overly Explicit Thematic Expression"}, and provide a structural optimization by inserting a narrative buildup node and adjusting causal relations to express the theme more implicitly.}\\
\vspace{0.1cm}
\texttt{\textbf{Problem Definition:}}\\
\texttt{\textbf{Overly Explicit Thematic Expression}: The theme is conveyed too directly through dialogue or narration, rather than being revealed progressively through character decisions, conflicts, and symbolic narrative elements.}\\
\vspace{0.1cm}
\texttt{\textbf{Optimization Guidelines:}}\\
\texttt{- Only analyze direct causal links in the relations field;}\\
\texttt{- Identify and remove overly abrupt or thematically obvious P → Z connections;}\\
\texttt{- Insert a buildup node Q between P and Z, where Q arises naturally from P and builds narrative tension;}\\
\texttt{- Output only one issue.}\\
\bottomrule
\caption{Prompts used by the Theme Agent for analyzing theme-related issues in the event graph.}
\label{prompt:theme_agent}
\end{longtable}
\end{small}

\begin{small}
\begin{longtable}{p{0.95\textwidth}}
\toprule
\textbf{Character Agent Analysis Prompts:} \\
\vspace{0.2cm}
\texttt{The Character Agent examines three aspects: (1) character drive, (2) character flatness, and (3) character arc abruptness.}\\
\vspace{0.1cm}
\textbf{1. Character Drive Analysis:}\\
\texttt{You are an expert in analyzing character psychological motivation. Please examine the following event graph to determine whether there exists an issue of \textbf{"Lack of Internal Motivation and Setup in Character Development"}, and suggest a structural optimization by inserting a motivation-building event node and reconstructing the causal chain.}\\
\vspace{0.1cm}
\texttt{\textbf{Problem Definition:}}\\
\texttt{\textbf{Lack of Internal Motivation and Setup}: The character makes a significant decision or undergoes a major attitude shift at a key plot point, but the preceding event lacks sufficient emotional response, internal reflection, or critical external stimulus to justify the change.}\\
\vspace{0.1cm}
\textbf{2. Character Flatness Analysis:}\\
\texttt{You are a character complexity design expert. Please analyze the following event graph to identify whether there is an issue of \textbf{"One-Dimensional Characterization"}, and suggest a structural optimization by inserting a fluctuation node and adjusting the causal event structure.}\\
\vspace{0.1cm}
\texttt{\textbf{Problem Definition:}}\\
\texttt{\textbf{One-Dimensional Characterization}: The character consistently behaves in a single manner, lacking emotional fluctuation, personality contrast, or internal conflict.}\\
\vspace{0.1cm}
\textbf{3. Character Arc Analysis:}\\
\texttt{You are a narrative pacing expert specialized in character arc development. Please analyze the following event graph to identify whether there is an issue of \textbf{"Abrupt Character Arc Shift"}, and provide a structural optimization suggestion by inserting a mediating node into an existing causal chain to better support the psychological transition of the character.}\\
\vspace{0.1cm}
\texttt{\textbf{Problem Definition:}}\\
\texttt{\textbf{Abrupt Character Arc Shift}: A character undergoes a significant behavioral or emotional change at a key narrative point, but the preceding event lacks sufficient setup in terms of motivation, emotion, conflict, or external stimulus.}\\
\bottomrule
\caption{Prompts used by the Character Agent for analyzing character-related issues in the event graph.}
\label{prompt:character_agent}
\end{longtable}
\end{small}

\begin{small}
\begin{longtable}{p{0.95\textwidth}}
\toprule
\textbf{Plot Agent Analysis Prompts:} \\
\vspace{0.2cm}
\texttt{The Plot Agent audits five structural dimensions: (1) plot incoherence, (2) missing turning points, (3) lack of foreshadowing, (4) insufficient suspense, and (5) relation conflicts.}\\
\vspace{0.1cm}
\textbf{1. Plot Incoherence Analysis:}\\
\texttt{You are a narrative progression structure optimization expert. Please analyze the following event graph to identify whether there is an issue of \textbf{"Incoherent Plot Progression"}, and propose a structural optimization by inserting a progression node to improve narrative continuity.}\\
\vspace{0.1cm}
\texttt{\textbf{Problem Definition:}}\\
\texttt{\textbf{Incoherent Plot Progression}: Adjacent events lack clear logical transitions or causal links, resulting in abrupt or unnatural narrative development.}\\
\vspace{0.1cm}
\textbf{2. Missing Suspense Analysis:}\\
\texttt{You are a suspense design and narrative pacing expert. Please analyze the following event graph to determine whether it contains an issue of \textbf{"Lack of Suspense"}, and propose a structural optimization by inserting a suspense node to establish a cross-phase tension chain.}\\
\vspace{0.1cm}
\texttt{\textbf{Problem Definition:}}\\
\texttt{\textbf{Lack of Suspense}: The plot reveals too much information too clearly and completely, leaving no room for mystery or open questions.}\\
\vspace{0.1cm}
\textbf{3. Lack of Foreshadowing Analysis:}\\
\texttt{You are a narrative structure optimization expert. Please analyze the following event graph to determine whether there is an issue of \textbf{"Lack of Foreshadowing"}, and provide a structural enhancement suggestion by embedding symbolic behaviors or implicit references in earlier events.}\\
\vspace{0.1cm}
\texttt{\textbf{Problem Definition:}}\\
\texttt{\textbf{Lack of Foreshadowing}: The script lacks symbolic details, hidden clues, or behavioral hints deliberately planted in early stages, which leads to an emotionally flat or structurally disconnected climax.}\\
\vspace{0.1cm}
\textbf{4. Plot Turning Point Analysis:}\\
\texttt{You are a narrative rhythm and dramatic structure expert. Please analyze the following event graph to identify whether there is an issue of \textbf{"Lack of Plot Reversal"}, and propose a structural optimization by inserting a reversal node to enhance dramatic variation.}\\
\vspace{0.1cm}
\textbf{5. Relation Conflict Analysis:}\\
\texttt{You are an expert in analyzing logical consistency in script relationships. Please assess the overall event graph to identify whether there is an issue of \textbf{"Contradictions in Character or Plot Relationships"}, and provide a suggestion for improving logical coherence.}\\
\bottomrule
\caption{Prompts used by the Plot Agent for analyzing plot-related issues in the event graph.}
\label{prompt:plot_agent}
\end{longtable}
\end{small}

\begin{small}
\begin{longtable}{p{0.95\textwidth}}
\toprule
\textbf{Constraint-Satisfaction Dialogue Generation Prompt:} \\
\texttt{You are generating \textbf{high-quality cinematic dialogue} that satisfies multiple narrative constraints.}\\
\vspace{0.1cm}
\texttt{\textbf{Scene Information:}}\\
\texttt{- \textbf{Location}: \{scene.place\}}\\
\texttt{- \textbf{Plot Element}: \{scene.plot\_element\}}\\
\texttt{- \textbf{Key Moment}: \{scene.beat\}}\\
\vspace{0.1cm}
\texttt{\textbf{Character Descriptions:}}\\
\texttt{\{json.dumps(characters, indent=2)\}}\\
\vspace{0.1cm}
\texttt{\textbf{Narrative Memory Context:}}\\
\texttt{\{memory\_summary\}}\\
\vspace{0.1cm}
\texttt{\textbf{Character State Constraints:}}\\
\texttt{\{json.dumps(character\_constraints, indent=2)\}}\\
\vspace{0.1cm}
\texttt{\textbf{Constraint Satisfaction Requirements:}}\\
\texttt{1. \textbf{Character Consistency Constraint}: Each character's dialogue must align with their description and current emotional state from memory.}\\
\texttt{2. \textbf{Narrative Coherence Constraint}: Dialogue must maintain logical flow with previous scenes via memory context.}\\
\texttt{3. \textbf{Character Arc Constraint}: Dialogue should reflect character development indicated in the narrative graph.}\\
\texttt{4. \textbf{Emotional Continuity Constraint}: Emotional trajectory from memory must be respected and advanced appropriately.}\\
\texttt{5. \textbf{Inter-scene Reference Constraint}: Naturally incorporate references to previous scenes when relevant.}\\
\vspace{0.1cm}
\texttt{\textbf{Dialogue Generation Guidelines:}}\\
\texttt{- Each character must speak at least once, reflecting their current state and constraints.}\\
\texttt{- Introduce at least one emotional shift that respects the character arc.}\\
\texttt{- Maintain logical coherence with previous dialogue and character arcs.}\\
\texttt{- Make each character's voice distinct and natural.}\\
\texttt{- Use emotion and subtext (hidden intentions, suppressed feelings) to add depth.}\\
\texttt{- Provide at least 8 dialogue turns (approx. 12–16 lines in total).}\\
\texttt{- Ensure all constraints are satisfied simultaneously.}\\
\vspace{0.2cm}
\texttt{Strictly return JSON format:}\\
\texttt{\{}\\
\texttt{  "dialogue": [}\\
\texttt{    "Character1: (emotion, action) Line...",}\\
\texttt{    "Character2: (reaction, subtext) Response...",}\\
\texttt{    ...}\\
\texttt{  ]}\\
\texttt{\}}\\
\bottomrule
\caption{Prompt for generating dialogue with constraint satisfaction framework incorporating narrative memory and character state constraints.}
\label{prompt:dialogue_constraint}
\end{longtable}
\end{small}

\begin{small}
\begin{longtable}{p{0.95\textwidth}}
\toprule
\textbf{Pairwise Script Evaluation Prompt (Storyline Comparison):} \\
\texttt{You are a professional script analyst. Please act as an impartial judge and evaluate the quality of the two scripts generated by different methods.}\\
\vspace{0.2cm}
\texttt{Your evaluation should focus \textbf{only} on the following dimension:}\\
\vspace{0.2cm}
\texttt{\textbf{Dimension – [Narrative/Thematic Expression/Characterization/Dramatic Engagement/Premise Fidelity]}}\\
\texttt{Evaluate based on the criteria specified in Table~\ref{tab:evaluation_dimensions}.}\\
\vspace{0.2cm}
\texttt{Evaluation Guidelines:}\\
\texttt{1. Avoid position biases: the order of presentation should not influence your decision.}\\
\texttt{2. Ignore superficial factors: do not let length, formatting style, or surface polish bias your judgment.}\\
\texttt{3. Focus on content quality: base your reasoning strictly on the narrative quality under the specified dimension.}\\
\texttt{4. Comparative assessment: compare the two scripts directly on the given dimension.}\\
\vspace{0.2cm}
\texttt{Decision Criteria:}\\
\texttt{• Choose "A" if Script A demonstrates clearly superior performance on this dimension.}\\
\texttt{• Choose "B" if Script B demonstrates clearly superior performance on this dimension.}\\
\texttt{• Choose "Same" if both scripts are approximately equal in quality, or if neither shows a significant advantage.}\\
\vspace{0.2cm}
\texttt{Input Format:}\\
\texttt{[Script A]}\\
\texttt{Premise A: \{premise\_a\}}\\
\texttt{Title A: \{title\_a\}}\\
\texttt{Summary (Beats):}\\
\texttt{\{beats\_a\}}\\
\vspace{0.2cm}
\texttt{[Script B]}\\
\texttt{Premise B: \{premise\_b\}}\\
\texttt{Title B: \{title\_b\}}\\
\texttt{Summary (Beats):}\\
\texttt{\{beats\_b\}}\\
\vspace{0.2cm}
\texttt{Output Requirements:}\\
\texttt{1. Provide a concise, one-sentence explanation justifying your judgment.}\\
\texttt{2. Output your verdict in strict JSON format as specified below.}\\
\vspace{0.2cm}
\texttt{Required JSON Format:}\\
\texttt{\{}\\
\texttt{  "explanation": "your explanation of which script is better and why",}\\
\texttt{  "verdict": "A" or "B" or "Same"}\\
\texttt{\}}\\
\bottomrule
\caption{Prompt template for pairwise comparison of storylines (beats only) between two scripts. The dimension placeholder is replaced with one of the five evaluation dimensions (see Table~\ref{tab:evaluation_dimensions}).}
\label{prompt:evaluation_storyline}
\end{longtable}
\end{small}

\begin{small}
\begin{longtable}{p{0.95\textwidth}}
\toprule
\textbf{Pairwise Script Evaluation Prompt (Full Script Comparison):} \\
\texttt{You are a professional script analyst. Please act as an impartial judge and evaluate the quality of the two scripts generated by different methods.}\\
\vspace{0.2cm}
\texttt{Your evaluation should focus \textbf{only} on the following dimension:}\\
\vspace{0.2cm}
\texttt{\textbf{Dimension – [Narrative/Thematic Expression/Characterization/Dramatic Engagement/Premise Fidelity]}}\\
\texttt{Evaluate based on the criteria specified in Table~\ref{tab:evaluation_dimensions}.}\\
\vspace{0.2cm}
\texttt{Evaluation Guidelines:}\\
\texttt{1. Avoid position biases: the order of presentation should not influence your decision.}\\
\texttt{2. Ignore superficial factors: do not let length, formatting, or surface polish affect your judgment.}\\
\texttt{3. Focus on content quality: base your reasoning strictly on the narrative quality under the specified dimension.}\\
\texttt{4. Comparative assessment: compare the two scripts directly on the given dimension.}\\
\vspace{0.2cm}
\texttt{Decision Criteria:}\\
\texttt{• Choose "A" if Script A demonstrates clearly superior performance on this dimension.}\\
\texttt{• Choose "B" if Script B demonstrates clearly superior performance on this dimension.}\\
\texttt{• Choose "Same" if both scripts are approximately equal in quality, or if neither shows a significant advantage.}\\
\vspace{0.2cm}
\texttt{Input Format:}\\
\texttt{[Script A]}\\
\texttt{Title: \{title\_a\}}\\
\texttt{Premise: \{premise\_a\}}\\
\texttt{Full Script:}\\
\texttt{\{scenes\_a\}}\\
\texttt{(Each scene includes: Place, Plot element, Beat, and Dialogue)}\\
\vspace{0.2cm}
\texttt{[Script B]}\\
\texttt{Title: \{title\_b\}}\\
\texttt{Premise: \{premise\_b\}}\\
\texttt{Full Script:}\\
\texttt{\{scenes\_b\}}\\
\texttt{(Each scene includes: Place, Plot element, Beat, and Dialogue)}\\
\vspace{0.2cm}
\texttt{Output Requirements:}\\
\texttt{1. Provide a concise, one-sentence explanation justifying your judgment.}\\
\texttt{2. Output your verdict in strict JSON format as specified below.}\\
\vspace{0.2cm}
\texttt{Required JSON Format:}\\
\texttt{\{}\\
\texttt{  "explanation": "your explanation of which script is better and why",}\\
\texttt{  "verdict": "A" or "B" or "Same"}\\
\texttt{\}}\\
\bottomrule
\caption{Prompt template for pairwise comparison of full scripts (including scenes and dialogue) between two scripts. The dimension placeholder is replaced with one of the five evaluation dimensions (see Table~\ref{tab:evaluation_dimensions}).}
\label{prompt:evaluation_full_script}
\end{longtable}
\end{small}

\begin{small}
\begin{longtable}{p{0.95\textwidth}}
\toprule
\textbf{Evaluation Dimensions and Criteria:} \\
\vspace{0.2cm}
\textbf{1. Narrative}\\
\texttt{Assess the narrative quality based on the following criteria:}\\
\texttt{• \textbf{Plot Continuity}: Smooth transitions between events with clear causal linkages}\\
\texttt{• \textbf{Logical Consistency}: Coherent contextual setups, world-building, storyline progression, and character behaviors}\\
\texttt{• \textbf{Dramatic Structure}: Presence of a complete narrative arc (Exposition, Rising Action, Climax, Falling Action, Resolution)}\\
\vspace{0.2cm}
\textbf{2. Thematic Expression}\\
\texttt{Assess the thematic development based on the following criteria:}\\
\texttt{• \textbf{Theme Clarity}: Clear and consistent central theme throughout the script}\\
\texttt{• \textbf{Theme Depth}: Sophisticated exploration of the theme with nuanced treatment}\\
\texttt{• \textbf{Artistic Reinforcement}: Effective use of metaphor, symbolism, and narrative devices to enrich thematic content}\\
\vspace{0.2cm}
\textbf{3. Characterization}\\
\texttt{Assess character portrayal based on the following criteria:}\\
\texttt{• \textbf{Motivation Credibility}: Clear and believable character motivations that drive actions}\\
\texttt{• \textbf{Character Depth}: Emotional and psychological complexity creating well-rounded, multi-dimensional characters}\\
\texttt{• \textbf{Character Development}: Evident growth or meaningful transformation with natural, well-paced development arcs}\\
\vspace{0.2cm}
\textbf{4. Dramatic Engagement}\\
\texttt{Assess dramatic tension and audience engagement based on the following criteria:}\\
\texttt{• \textbf{Event Design}: Well-crafted, compelling events that sustain audience interest}\\
\texttt{• \textbf{Suspense Construction}: Effective use of foreshadowing, hints, and delayed revelations}\\
\texttt{• \textbf{Narrative Pacing}: Significant turning points that shift stakes or character trajectories}\\
\texttt{• \textbf{Tension Management}: Skillful buildup and release of dramatic tension throughout the narrative}\\
\vspace{0.2cm}
\textbf{5. Premise Fidelity}\\
\texttt{Assess adherence to the original premise based on the following criteria:}\\
\texttt{• \textbf{Conceptual Fidelity}: Faithful adherence to the core idea and thematic direction of the given premise}\\
\texttt{• \textbf{Element Retention}: Core premise elements—primary settings, characters, and central conflicts—are faithfully retained}\\
\bottomrule
\caption{The five evaluation dimensions and their specific criteria used for pairwise script comparison. Each dimension is assessed independently, and the evaluator compares two scripts directly on each dimension's criteria.}
\label{tab:evaluation_dimensions}
\end{longtable}
\end{small}

\begin{small}
\begin{longtable}{p{0.95\textwidth}}
\toprule
\textbf{Example: Complete Evaluation Prompt for Narrative Dimension (Storyline):} \\
\texttt{You are a professional script analyst. Please act as an impartial judge and evaluate the quality of the two scripts generated by different methods.}\\
\vspace{0.2cm}
\texttt{Your evaluation should focus \textbf{only} on the following dimension:}\\
\vspace{0.2cm}
\texttt{\textbf{Dimension – Narrative}}\\
\texttt{Evaluate the narrative based on the following criteria:}\\
\texttt{• Plot continuity and smooth transitions between events}\\
\texttt{• Logical consistency in contextual setups, world-building, storyline, and character behaviors}\\
\texttt{• Presence of a clear and complete dramatic narrative structure (typically encompassing Exposition, Rising Action, Climax, Falling Action, Resolution)}\\
\vspace{0.2cm}
\texttt{Avoid any position biases and ensure that the order in which the scripts are presented does not influence your decision. Do not let the length or formatting style of the summaries bias your judgment. Base your reasoning strictly on the content quality under the specified evaluation dimension.}\\
\vspace{0.2cm}
\texttt{Choose "A" if Script A clearly demonstrates superior performance on this dimension.}\\
\texttt{Choose "B" if Script B clearly demonstrates superior performance on this dimension.}\\
\texttt{Choose "Same" if both scripts are roughly equal in quality on this dimension, or if neither shows a significant advantage.}\\
\vspace{0.2cm}
\texttt{After comparing the two summaries, provide a one-sentence explanation of your judgment, and then output your final verdict in strict JSON format.}\\
\vspace{0.2cm}
\texttt{[Script A]}\\
\texttt{Premise A: [premise text]}\\
\texttt{Title A: [title]}\\
\texttt{Summary:}\\
\texttt{[beat summaries from all scenes]}\\
\vspace{0.2cm}
\texttt{[Script B]}\\
\texttt{Premise B: [premise text]}\\
\texttt{Title B: [title]}\\
\texttt{Summary:}\\
\texttt{[beat summaries from all scenes]}\\
\vspace{0.2cm}
\texttt{Respond strictly in the following JSON format:}\\
\texttt{\{}\\
\texttt{  "explanation": "your explanation of which script is better and why",}\\
\texttt{  "verdict": "A" or "B" or "Same"}\\
\texttt{\}}\\
\bottomrule
\caption{Complete example of the evaluation prompt for the Narrative dimension (storyline comparison). The same template structure is used for all five dimensions, with the dimension-specific criteria replaced accordingly.}
\label{prompt:evaluation_example}
\end{longtable}
\end{small}

\twocolumn

\end{document}